# Learning Mechanism Underlying NLP Pre-Training and Fine-Tuning

**Yarden Tzach[a,1], Ronit D. Gross[a,1], Ella Koresh[a,1], Shalom Rosner[b], Or Shpringer[b], Tal Halevi[a] and Ido Kanter[a,b,*]**

[a]Department of Physics, Bar-Ilan University, Ramat-Gan, 52900, Israel.

[b]Gonda Interdisciplinary Brain Research Center, Bar-Ilan University, Ramat-Gan, 52900, Israel.

*Corresponding author at: Department of Physics, Bar-Ilan University, Ramat-Gan, 52900, Israel. E-mail address: ido.kanter@biu.ac.il (I. Kanter).

[1]These authors equally contributed to this work

## Abstract

Natural language processing (NLP) enables the understanding and generation of meaningful human language, typically using a pre-trained complex architecture on a large dataset to learn the language and next fine-tune its weights to implement a specific task. Twofold goals are examined; to understand the mechanism underlying successful pre-training and to determine the interplay between the pre-training accuracy and the fine-tuning of classification tasks. The following main results were obtained; the accuracy per token (APT) increased with its appearance frequency in the dataset, and its average over all tokens served as an order parameter to quantify pre-training success, which increased along the transformer blocks. Pre-training broke the symmetry among tokens and grouped them into finite, small, strong match token clusters, as inferred from the presented token confusion matrix. This feature was sharpened along the transformer blocks toward the output layer, enhancing its performance considerably compared with that of the embedding layer. Consequently, higher-order language structures were generated by pre-training, even though the learning cost function was directed solely at identifying a single token. These pre-training findings were reflected by the improved fine-tuning accuracy along the transformer blocks. Additionally, the output label prediction confidence was found to be independent of the average input APT, as the input meaning was preserved since the tokens are replaced primarily by strong match tokens. Finally, although pre-training is commonly absent in image classification tasks, its underlying mechanism is similar to that used in fine-tuning NLP classification tasks, hinting at its universality. The results were based on the BERT-6 architecture pre-trained on the Wikipedia dataset and fine-tuned on the FewRel and DBpedia classification tasks.

## 1. Introduction

Natural language processing (NLP) is currently a rapidly developing research field with a broad range of applications. Moreover, its implementation typically requires two consecutive learning processes, pre-training and fine-tuning[1-5].

The pre-training process aims to embed natural language in a given architecture using a large dataset, such as the Wikipedia dataset[6], which exemplifies the complexity of the language structures. However, it requires enormous computational resources as the training dataset can exceed terabytes in size, and complex architectures can exceed billions of tuned parameters[7].

The language space can be defined using the following dimensions: The architecture output typically comprises a few tens of thousands of tokens ($T_{Number}$) representing the fundamental richness of the language. For example, each token can represent a word, part of a word, or a punctuation mark, depending on the tokenizer details[8, 9]. An input comprises a fixed token sequence length ($N_{Input}$) based on a sentence or several sentences, and is padded by a special token if it is shorter than $N_{Input}$. Finally, each token is embedded into a vector of rank $E_{Length}$, which is tuned during the pre-training process. Consequently, the language is represented by a tunable $(T_{Number}, E_{Length})$ matrix and an input is represented by a $(N_{Input}, E_{Length})$ matrix.

There are several ways to realize the pre-training process. The common method being token masking[10], whereby a small finite fraction $(15\%)$ of random tokens is selected from the input token sequence in each iteration, most of which $(80\%)$ are erased and replaced by the mask token, half of the remaining selected tokens are replaced by random tokens, and the rest are unchanged. The architecture must then determine the most probable selected token in each token masking location for each sequence element and must be trained accordingly. A perfect system correctly guesses all selected tokens.

The pre-training weights serve as the initial weights for solving a specific task, such as the classification of sentences (or paragraphs) into several categories ($N_{Labels}$). The output layer is reduced from $T_{Number}$ to $N_{Labels}$, and the architecture is then fine-tuned (transfer learning[11, 12]) using the classification training dataset, with its success estimated using the test accuracy. The pre-training process substantially improves accuracy compared to fine-tuning the same architecture using random initial weights.

The usefulness of pre-training in solving NLP tasks has been demonstrated for large language models (LLMs)[7]. However, very recently it was demonstrated also for tiny language models (TLMs) consisting of $\sim 10^{-3}$ smaller pre-training dataset sizes compared with those used for LLMs[5]; however, they cover a substantial fraction of the total token space[6, 7], reflecting their linguistic richness[8]. Results indicated that TLMs exhibit a clear performance gap between pre-trained and non-pre-trained models across classification tasks, demonstrating the effectiveness of pre-training, even at a tiny pre-training scale[5]. Nevertheless, the performance gap increases with the size of the pre-training dataset and with a greater overlap between tokens in the pre-training and classification datasets. These results have opened new avenues for researchers to deepen their understanding of the mechanism underlying NLP without the enormous computational resources required for LLM pre-training, which is currently feasible for only a few dominant companies.

This study has two primary goals. First, it aims to understand the mechanism underlying successful pre-training. Specifically, to find an order parameter capable of quantifying the pre-training success along the architecture's transformer blocks. Additionally, it aims to examine whether higher-order language structures can be learned beyond the identification of an individual token, the direct pre-training task, and compare it with the embedding layer ability. The second goal is to determine the interplay between the pre-training success and the fine-tuning classification accuracy along the architecture's transformers[5]. If this interplay is similar to that previously determined for image classification tasks using convolutional neural networks[13-16] and vision transformer architectures[17-20], it will indicate a universal mechanism underlying the classification tasks[21-23].

The results presented have been derived from simulations using either BERT-6[24] pre-trained on the entire Wikipedia dataset comprising approximately six million paragraphs[9] or BERT-6 pre-trained on a tiny random Wikipedia subset of size $W_S \epsilon [40{,}000, 90{,}000]$[10], where $E_{Length} = 768$ and $N_{Input} = 128$. The accuracy of the subsequent fine-tuning was measured for the FewRel[11] and DBpedia[13,14] classification tasks[15].

The main results demonstrate that the order parameter to quantify pre-training success are the average APT, the average probability that a mask token has been correctly predicted. This probability is inhomogeneous among the tokens and increases on average with the frequency of the token occurrence in the pre-trained dataset. Additionally, the average accuracy per token increases with $W_S$ and along the transformer blocks toward the output layer, indicating enhanced learning along the deep architecture.

The most striking result is that pre-training breaks the local symmetry among tokens, grouping them into strongly matched finite small clusters, as indicated by the token confusion matrix. When the predicted masked token is incorrect, it has a high probability of having a similar meaning. The aggregation of a token with a group of tokens that have similar meanings or functionalities is a strong indication of the generation of higher-order language structures by pre-training, even though the cost function of training is based solely on identifying a single token.

This local symmetry breaking among tokens is generalized to cluster formation among tokens, similar to percolation clusters[25, 26]. This type of local symmetry breaking (small clusters) is sharpened along the transformer blocks towards the output layer, demonstrating enhanced language learning in comparison with the embedding layer's performance[27, 28]. This is reflected in the enhanced fine-tuning accuracy along the transformer blocks. In addition, the confidence in predicting the output label is independent of the average accuracy of the tokens composing its input.

Finally, although pre-training is typically absent in image classification tasks, its underlying mechanism is similar to that of fine-tuning NLP classification tasks, hinting at its universal behavior.

## 2. Results

### 2.1. Accuracy per token (APT)

#### 2.1.1. Pre-training the entire Wikipedia dataset

A randomly selected test dataset of size $W^{test} = 90{,}000$ was used after the pre-training process over the entire Wikipedia dataset. During the testing phase, tokens were randomly selected from each input and modified (masked, replaced, or unchanged), and the architecture was trained to identify the original tokens. In such a relatively low-complexity pre-training realization $\sim 29{,}000$ tokens appeared in $W^{test}$, almost all $30{,}522$ tokens of the entire Wikipedia dataset. The accuracy per token (APT) was estimated as the ratio between the number of correct predictions of a given selected token in $W^{test}$ and the total number of times it was selected to be modified (masked/replaced/unchanged). To improve APT statistics for tokens that are less frequent in $W^{test}$, the test dataset was presented several times, typically $30$ repetitions, and the average APT over all tokens in $W^{test}$ is defined as <APT>.

The results indicated large fluctuations among the APT (Fig. 1). However, the average APT over token groups with similar appearance frequency increases with the frequency (Fig. 1). Nevertheless, the average APT over all tokens, the global quantity <APT>, is the fundamental order parameter to estimate the quality of the pre-training process.

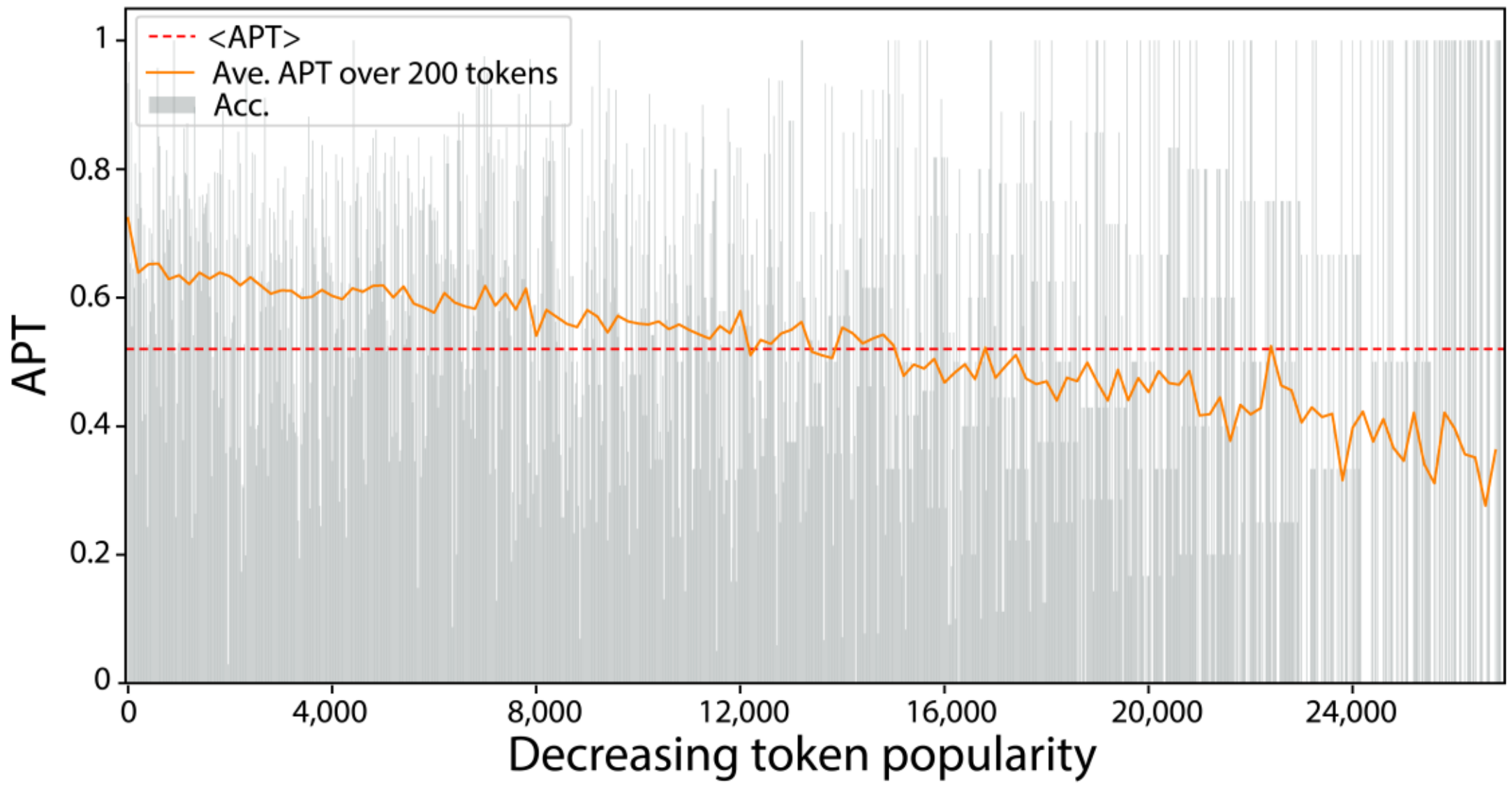

**Fig 1.** APT of each individual token ordered from most popular token to least popular (Grey) pre-trained on the entire Wikipedia dataset on BERT-6 and <APT> (Horizontal dashed-red line). Average APT over $200$ non-overlapping groups of consecutive tokens (Orange).

### 2.1.2. Tiny pre-training dataset

Pre-training the entire Wikipedia dataset is a heavy computational task; however, similar results and trends were found for tiny pre-training datasets consisting of $W_S = 90{,}000$ randomly selected Wikipedia paragraphs.

The quality along the six transformer blocks of the pre-trained BERT-6 was quantified using the following procedure, where the first $m(\leq 6)$ transformer blocks were kept unchanged (frozen). The $768$ output units of the $m^{th}$ block—representing pre-training by the partial architecture, were fully connected (FC) to $T_{Number}$ output units of the output layer, representing the tokens. This FC layer was trained to minimize the loss function. Finally, the APT and <APT> were estimated using $W^{test}$ dataset, indicating a progressive increase of <APT> with $m$ (Table 1). This indicates a pre-training enhancement along the deep architecture from embedding to output (Table 1). In addition, tokens with high APT approach their asymptotic learning values much faster along the architecture's blocks, as indicated by the average APT of the tokens with the $50$ highest APT (Table 1); hence, the significant contribution of this improvement is attributed to tokens with low APT.

| Block | 6 | 5 | 4 | 3 | 2 | 1 | 0 |
|---|---|---|---|---|---|---|---|
| <APT> | 0.31 | 0.284 | 0.227 | 0.202 | 0.17 | 0.085 | 0.037 |
| APT(50) | 0.731 | 0.715 | 0.682 | 0.628 | 0.585 | 0.366 | 0.123 |

**Table 1.** The progressive <APT> and the average $50$ highest APT, APT(50), along the transformer blocks for a pre-trained BERT-6 using $W_S = 90{,}000$ and $W^{test} = 90{,}000$.

Another expected trend is that the APT increases with $W_S$ (Fig. 2) because the frequency of each token increases and its multiple language functionalities and embedding structures are better sampled.

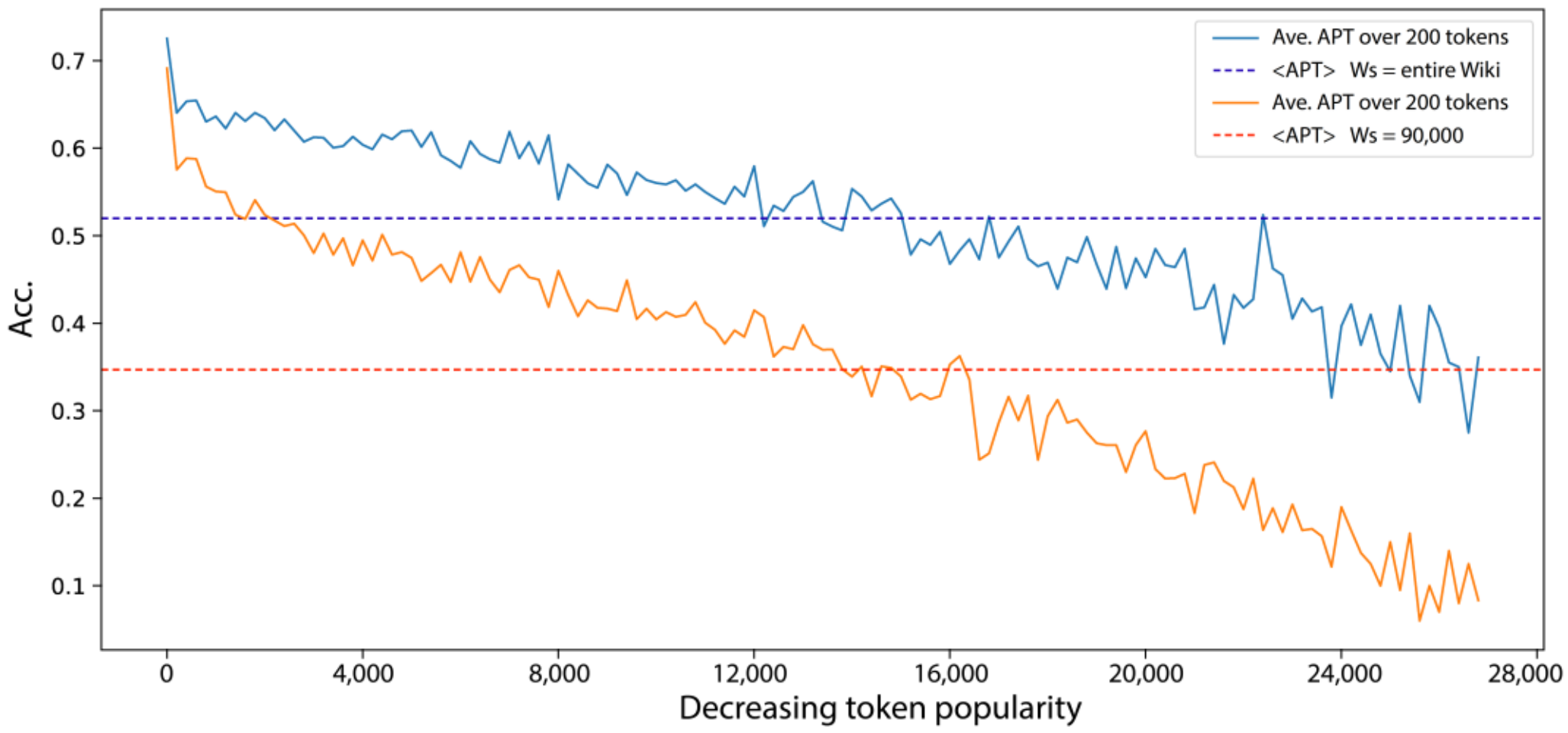

**Fig. 2.** Average APT over $200$ non-overlapping groups of consecutive tokens ordered from most popular tokens to least popular, pre-trained on the entire Wikipedia dataset and tested on $W^{test} = 90{,}000$ (Blue), and <APT> (Horizontal blue-dashed line). Same as Blue but pre-trained on $W_S = 90{,}000$ (Orange), and <APT> (Horizontal, orange-dashed line).

### 2.2. Confusion matrix and high-order language structures

An insight into the quality of the pre-training process beyond the APT is provided by the confusion square matrix $M_{Confus}$ with rank $T_{Number}$, the number of tokens composing $W_S$. The $(i, j)$ element of $M_{Confus}$ represents the number of times a modified (masked/replaced/unchanged) token $i$ is identified as token $j$. In case $M_{Confus}$ is diagonal APT=<APT>=1. To reduce noise in the following measurements, tokens that were poorly learned, where the diagonal element was not maximal in their rows, were excluded.

To generate $M_{Confus}$ after pre-training, $W^{test}$ randomly selected inputs were presented with only one random modified token per input, and at least $30$ repetitions over $W^{test}$ were performed to enhance the statistics of infrequent tokens. The modified token was masked by 80%, replaced with another token by 10%, and remained unchanged by 10%. The condition of one modified token per input leads to the building of each $M_{Confus}$ row independently, with the payoff of its slow generation; however, it excludes dependencies among multiple rows. Nevertheless, the qualitative results reported below are valid for multiple modified tokens per input.

Each row of the confusion matrix $M_{Confus}$, was normalized by its diagonal element. The distribution of the off-diagonal $M_{Confus}$ elements indicates the following trend: Most elements were relatively small; however, there was a long discontinuous tail with an increasing number of appearances per bin (Fig. 3). These large off-diagonal elements represent the predominantly confused tokens with the selected tokens in the input. The binary confusion matrix, $M_{Binary\ Confus}$, is formed by setting the off-diagonal and diagonal elements above the threshold of $0.05$ (Fig. 3b) to one and the rest to zero.

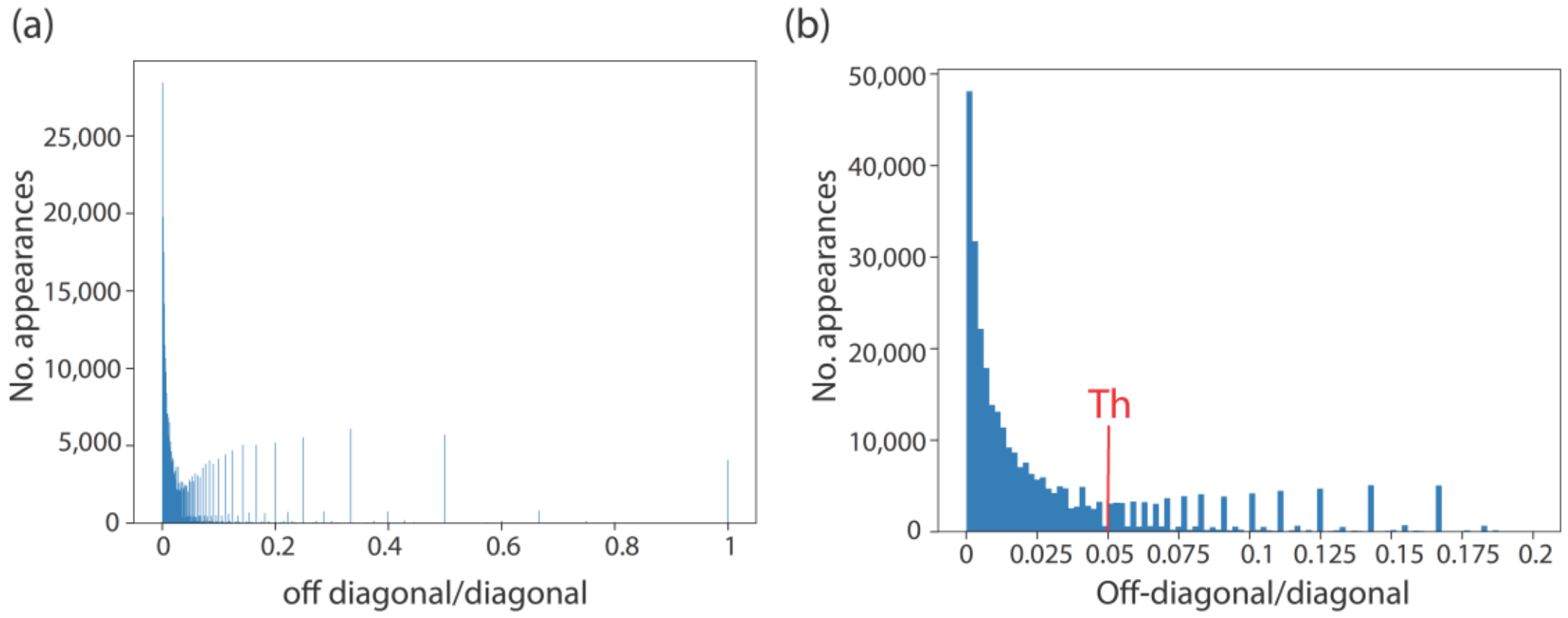

**Fig. 3.** (a) Distribution of the confusion matrix, $M_{Confus}$, non-zero off-diagonal elements. (b) Zoom-in on panel a in the range $[0, 0.2]$, where $Th = 0.05$ represents the threshold where the distribution of off-diagonal elements become notably sparser.

The product of all pairs of elements with the same indices $(i, j)$ in $M_{Binary\ Confus}$ and its transposed matrix

$$M_{adjacency}(i,j) = M_{Binary\ Confus}(i,j) \times M^{+}_{Binary\ Confus}(i,j) \quad (1)$$

result in an adjacency matrix representing a graph with symmetric connectivity between nodes (tokens). Token clusters were formed using a standard linear complexity percolation algorithm [26]. For each selected token, a cluster is formed by adding all the tokens that are connected to it. All tokens connected to the first token are added to the cluster, and the procedure is repeated for all newly added tokens. This procedure creates a percolating growth of all tokens through shared mutual connections. A cluster is formed when no additional tokens are added through the above threshold connections. The algorithm terminates when all the remaining tokens are attributed to a cluster. A perfect identification of tokens with no confusion consists of clusters of size $1$, whereas a meaningless one consists of one large cluster encompassing all. An intermediate structure consists of small clusters in addition to unity clusters, representing each ambiguity of the tokens belonging to the cluster. Tokens belonging to a cluster with strong similar language meanings (strongest match) represent low uncertainty, whereas a cluster with a weak match between its tokens indicates a worthy aggregation of tokens.

### 2.2.1 Confusion matrix using the entire Wikipedia dataset

For the pre-trained BERT-6 over the entire Wikipedia dataset, $M_{adjacency}$ was calculated using the $M_{Binary\ Confus}$ that was derived using $W^{test} = 90{,}000$ and $30$ repetitions. The list of cluster sizes (Table 2a) indicates $\sim 20{,}000$ clusters of size $1$, many small clusters below size $10$, and a few larger clusters with a maximal cluster size of $140$. Some of the small clusters are depicted (Fig. 4), where each group strong match tokens. For instance, different directions (Fig. 4b), types of travel paths (Fig. 4d), frequency of events (Fig. 4e), and opposite meanings (Fig. 4c). The exchange of a token in a sentence with another token belonging to its cluster is expected to be grammatically correct and, in many cases, to preserve the meaning. Nevertheless, the use of symmetric connections among tokens may deteriorate the strong match among distant tokens in the cluster, where the shortest path between them consists of a few intermediate tokens. For instance, the tokens of 'produce' and 'promote' (Fig.4f) and tokens of 'designed' and 'founded' (Fig.4a). This drift in the meaning of a token is expected as a result of match decay along a sequence of similar nearest-neighbor tokens.

The formation of larger clusters also indicates a common meaning among tokens. Small numbers below $25$ are grouped into a relatively dense cluster (Fig. 5a) as well as all the $12$ months (Fig.5b), where the token of 'can' is connected to the token of 'may', which has a double meaning in English, and the quality of the current pre-training cannot distinguish between both. The cluster of primarily Russian names (Fig. 5c) indicates that the pre-training identified their common source, although it could not uniquely define a name for a given sentence. Similarly, ordinal numbers until 22nd are clustered (Fig. 5d), where it is remarkable that the tokens of 'first', 'second', and 'third' for instance, are sequentially connected by a chain to the cluster, indicating their enhanced similarity in comparison to distant tokens in the cluster.

The many clusters of size $1$, $\sim 20{,}000$ (Table 2a), indicate that these tokens are uniquely defined up to their APT. Indeed, the average APT of these unity clusters, $0.614$, is found to be much higher than the average APT, $0.448$, of tokens belonging to clusters larger than unity (Fig. 6). Hence, supreme pre-training must approach the following two features: the fraction of tokens belonging to unity clusters approaches one as well as their average APT.

The results clearly demonstrate the efficiency of pre-training in segmentation of tokens into several meaningful clusters of various sizes. The generation of confusion and adjacency matrices and the resulting clusters present a method for estimating the efficiency of the pre-training process. Notably, the qualitatively reported results are insensitive to the exact threshold value in the proximity of $0.05$.

(a)

| Cluster Size | No. |
|---|---|
| 1 | 20,460 |
| 2 | 748 |
| 3 | 120 |
| 4 | 43 |
| 5 | 14 |
| 6 | 7 |
| 7 | 4 |
| 8 | 6 |
| 9 | 1 |
| 10 | 2 |
| 11 | 2 |
| 12,13,14, 25,30,33, 113,130,140 | 1 |

(b)

| Cluster Size | No. |
|---|---|
| 1 | 16,523 |
| 2 | 655 |
| 3 | 140 |
| 4 | 43 |
| 5 | 22 |
| 6 | 12 |
| 7 | 13 |
| 8 | 3 |
| 9 | 1 |
| 10,11 | 2 |
| 12 | 4 |
| 13 | 1 |
| 14 | 2 |
| 18,20,24, 110,276 | 1 |

(c)

| Cluster Size | No. |
|---|---|
| 1 | 13,216 |
| 2 | 534 |
| 3 | 87 |
| 4 | 47 |
| 5 | 18 |
| 6 | 11 |
| 7 | 8 |
| 8 | 5 |
| 9 | 1 |
| 10 | 2 |
| 11 | 1 |
| 12 | 2 |
| 15,16,19,21 | 1 |
| 25 | 2 |
| 62,136 | 1 |

**Table 2.** (a) Distribution of token cluster sizes for the 6th block of the BERT-6 architecture pre-trained on the entire Wikipedia dataset and tested on $W^{test} = 90{,}000$. (b) Same as panel a but pre-trained on $W_S = 90{,}000$ and tested on $W^{test} = 90{,}000$. (c) Same as panel b, but the distribution of token cluster sizes

was extracted for the 2nd block of BERT-6. Note that non-maximal diagonal tokens in the confusion matrix were excluded.

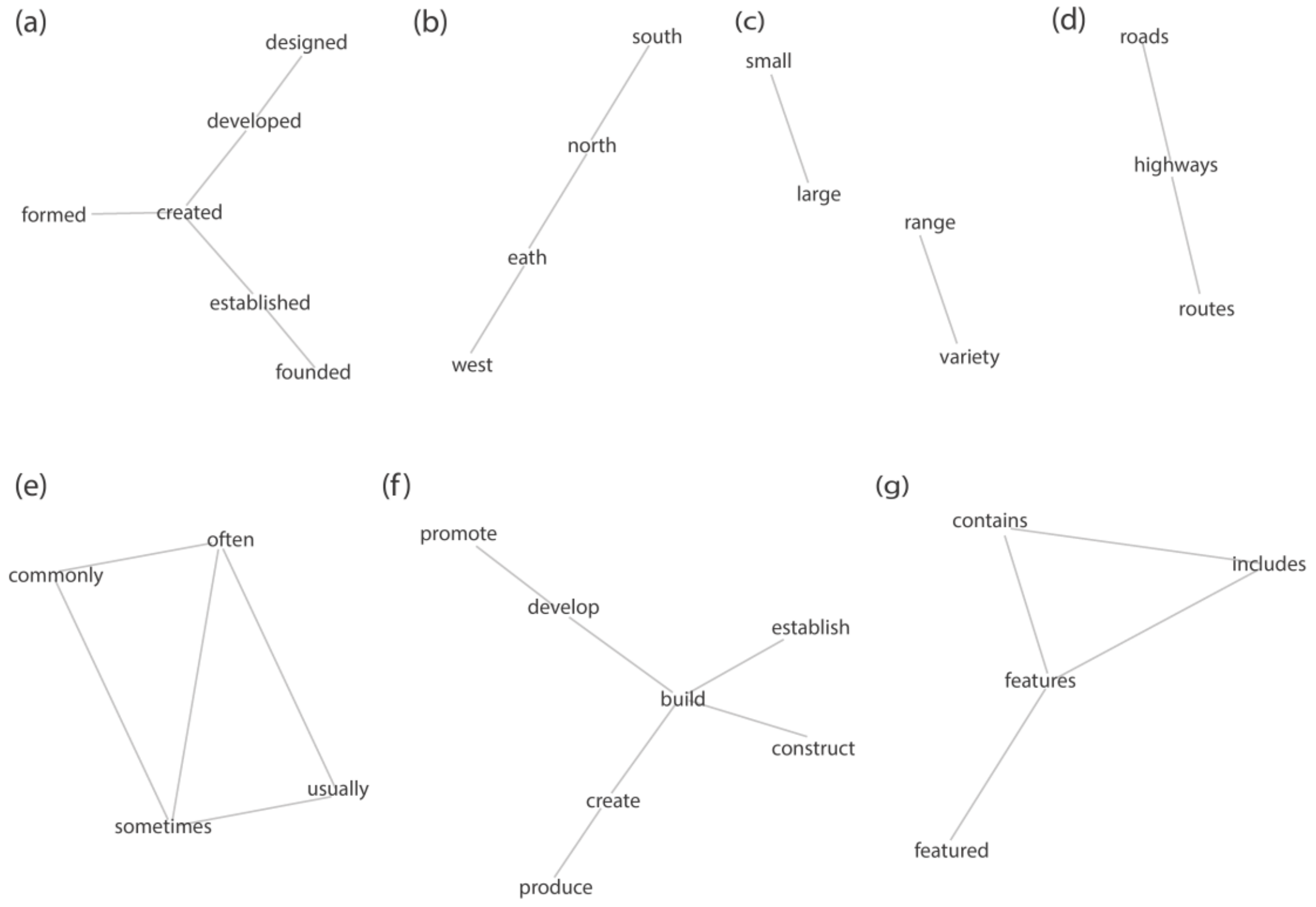

**Fig. 4.** Exemplified notable small clusters generated by the 6th block of BERT-6 architecture pre-trained on the entire Wikipedia and tested on $W^{test} = 90{,}000$.

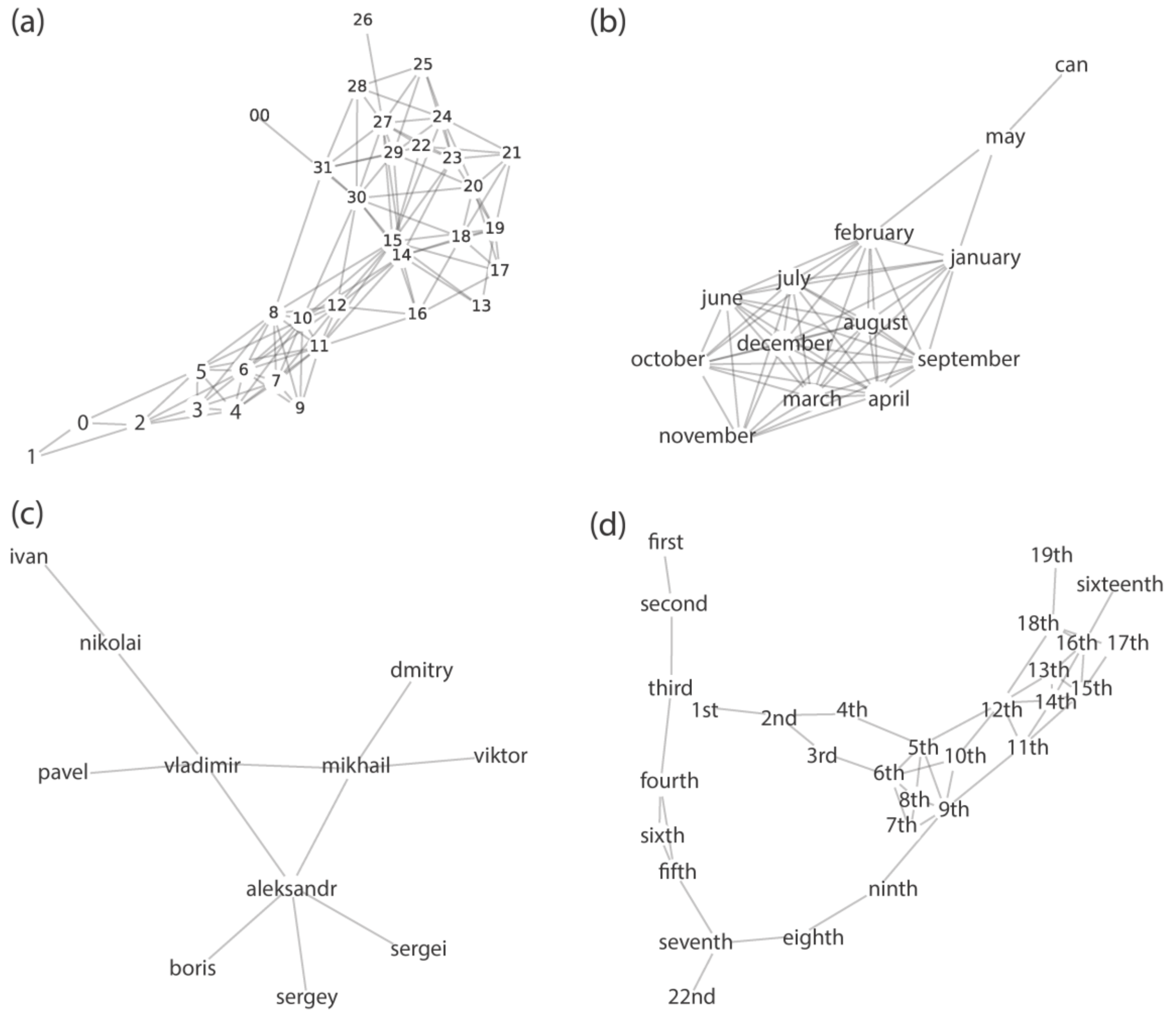

**Fig. 5.** Exemplified notable large clusters generated by the 6th block of BERT-6 architecture pre-trained on the entire Wikipedia and tested on $W^{test} = 90{,}000$.

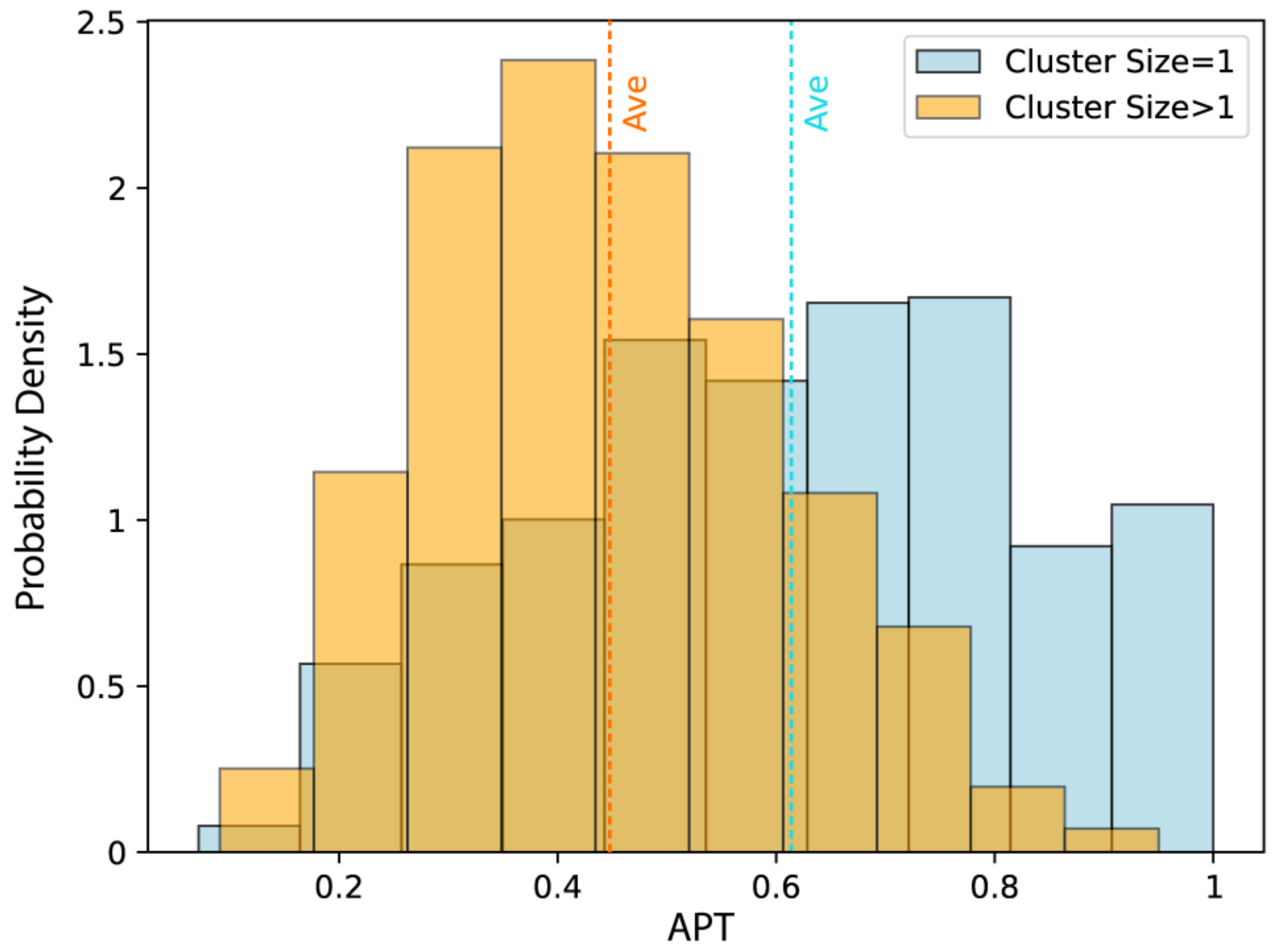

**Fig. 6.** Distribution of APT for tokens appearing in clusters of size 1 (light blue) with average $0.614$ (Vertical dashed-blue line) and for cluster sizes greater than 1 (Orange) with average $0.448$ (Vertical dashed-orange line). The cluster sizes were obtained for BERT-6 pre-trained on the entire Wikipedia dataset and $W^{test} = 90{,}000$.

### 2.2.2 Confusion matrix using tiny pre-train dataset

The robustness of meaningful cluster formation by pre-training over all Wikipedia datasets is extended to pre-training BERT-6 over a tiny dataset consisting of $W_S = 90{,}000$ randomly selected Wikipedia paragraphs only. Because the computational complexity of such pre-training is significantly reduced, it enables a comparison of the pre-processing progress along the transformer blocks.

The list of clusters and their sizes (Table 2b) is similar to the case of pre-training over the entire Wikipedia dataset; however, the number of clusters of size one decreased to $\sim 16{,}000$. Here again, the trend is preserved such that the average accuracy of tokens belonging to the unit clusters, $0.483$, is higher than that of the remaining tokens, $0.378$, where tokens with non-maximal diagonal elements were excluded. As expected, these accuracies were below the accuracies obtained for pre-training over the entire Wikipedia dataset.

The small clusters (Fig. 7) are similar to those obtained from pre-training over the entire Wikipedia dataset (Fig. 4), for instance, directions (Fig. 7b) and opposite meaning (Fig. 7c), where Fig. 7d is exceptional, including two nouns, 'roads' and 'transport', which are closely related in their meaning but cannot be identified as strong-match words with slightly different meanings. The larger clusters are also similar (Fig. 8) to those obtained for the entire Wikipedia (Fig. 5), where the $12$ months are grouped together without the anomaly of the token 'can' connected to 'may' (Fig. 8b), intermediate directions (Fig. 8a), and frequency of events (Fig. 8c) which is now much larger than the similar one (Fig. 4e) obtained for the entire Wikipedia.

The clusters obtained from Wikipedia are not expected to be exactly the same as those for a small $W_S$, where fluctuations between different samples are evident. Nevertheless, the high cluster similarity indicates that pre-training analysis can be performed using small datasets.

The pre-training quality of the 2nd transformer block of BERT-6 was quantified using the following procedure, in which the first two transformer blocks were kept unchanged (frozen). The $768$ output units of the $2^{nd}$ block, representing the pre-training by the partial architecture, were fully-connected (FC) to the output layer, with $T_{Number}$ output units representing the tokens. This FC layer was trained to minimize the loss function. Finally, $M_{Confus}$ and the binary confusion matrix $M_{Binary\ Confus}$ were formed. Using $M_{adjacency}$ (Eq. 1), the number and size of the clusters were obtained (Table 2c), and small and large clusters were observed (Figs. 9 and 10), similar to those obtained for the output of the 6th transformer block (Figs. 7 and 8).

The number of clusters of size one decreased to $\sim 13{,}000$ (Table 2c) in comparison to $\sim 16{,}000$ in the 6th transformer block (Table 2b), and the gap between the average accuracy of tokens of size one $(0.327)$ and the others $(0.285)$, decreased to $0.04$. Both results indicate a decrease in the quality of the 2nd block pre-training compared with the last, which is also supported by the exemplified small and large clusters (Figs. 9 and 10). The small clusters contain weakly matched tokens or a partial list of similar tokens. For instance, tokens of '#' and 'set' (Fig. 9a), 'perth' (Fig. 9g), and only four seemingly random English letters (Fig. 9e). In the large cluster of the 12 months 'would' is added using a connection to 'can' (Fig. 10b), the large numbers '100' and '200' were added to the cluster of selected small numbers (Fig. 10d), 'ferdinand' and 'maximillian' as well as other French, German, and Spanish names, are added to many American names (Fig. 10c), and 'recorded' seems to be a weak match to other cluster tokens (Fig. 10a), where in addition there is typically a stronger match between nearest-neighbor tokens and a possible drift meaning for tokens separated by several connections. These results indicate that higher-order language structures are enhanced and become more accurate during pre-training along the transformer blocks.

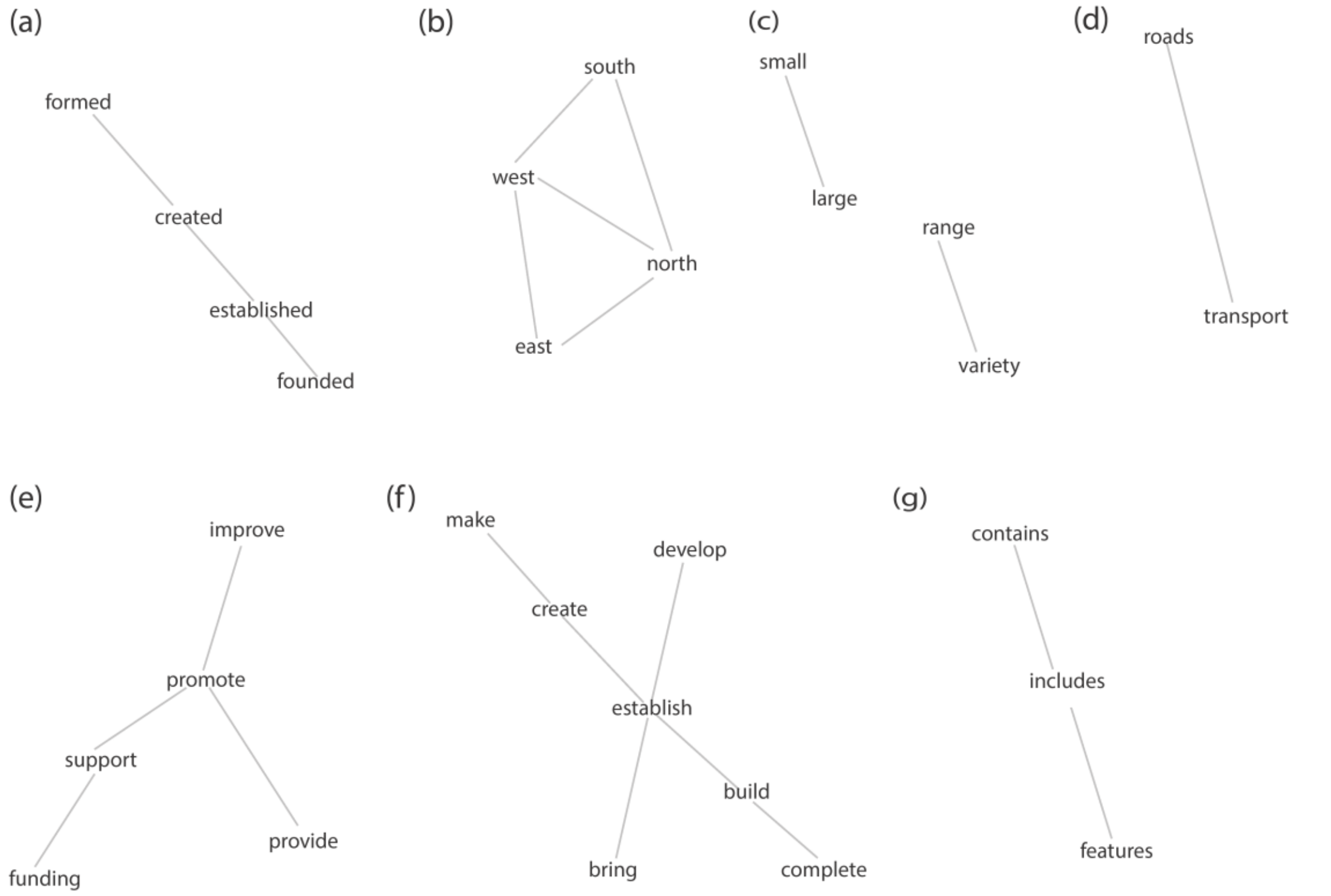

**Fig. 7.** Exemplified notable small clusters generated by the 6th block of BERT-6 architecture pre-trained and tested on $W_S = W^{test} = 90{,}000$.

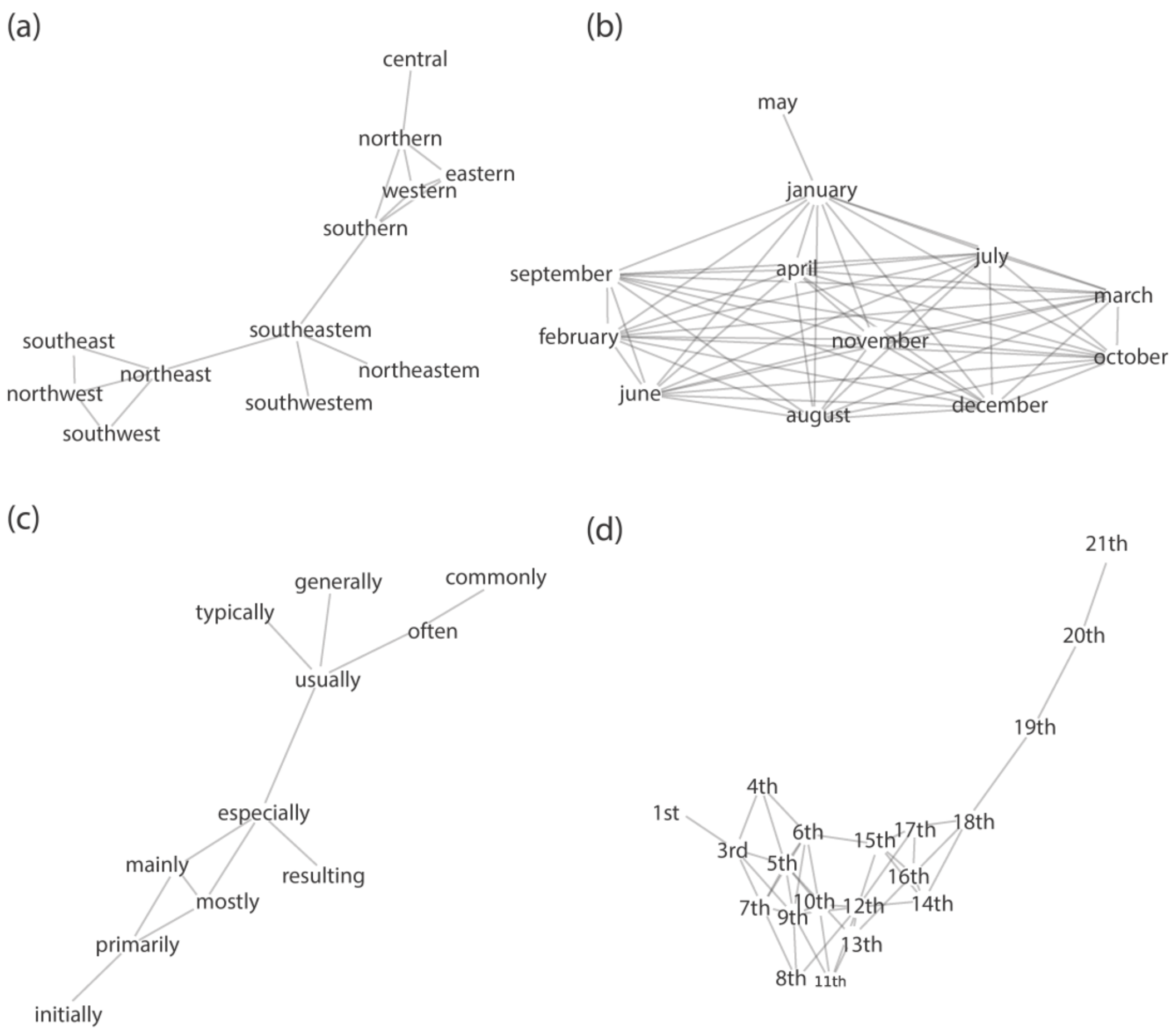

**Fig. 8.** Exemplified notable large clusters generated by the 6th block of BERT-6 architecture pre-trained and tested on $W_S = W^{test} = 90{,}000$.

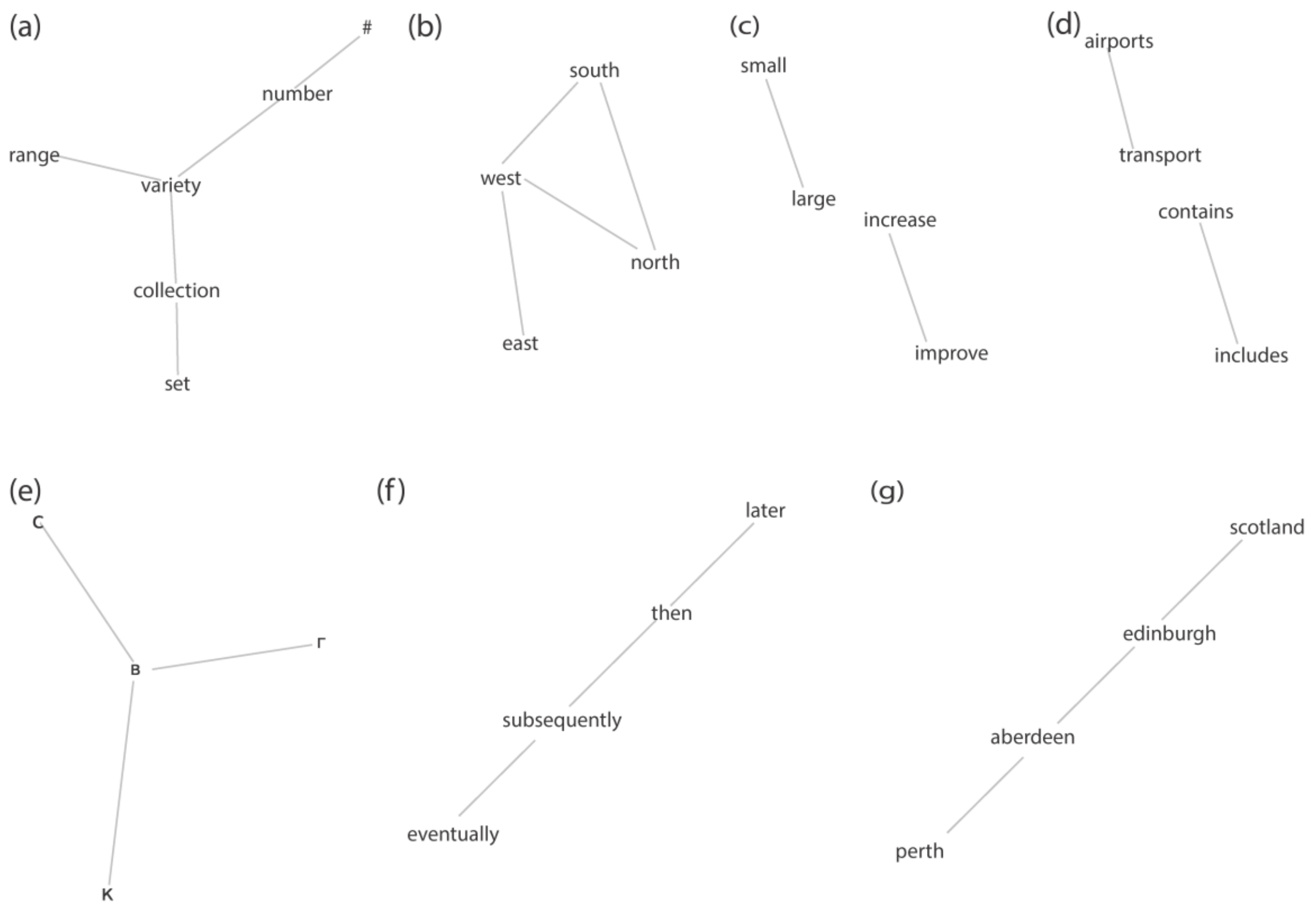

**Fig. 9.** Exemplified notable small clusters generated by the 2nd block of BERT-6 architecture pre-trained and tested on $W_S = W^{test} = 90{,}000$.

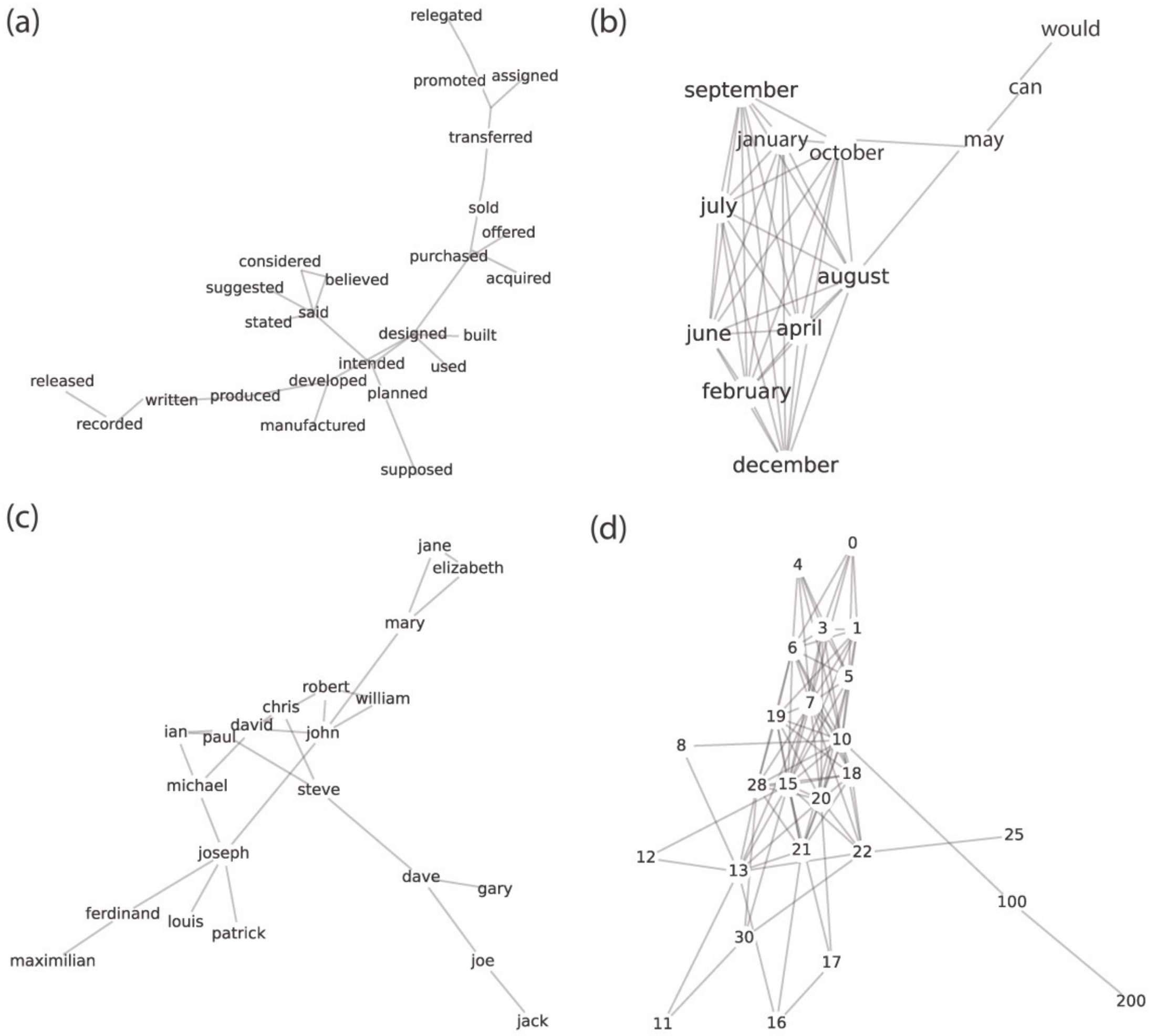

**Fig. 10.** Exemplified notable large clusters generated by the 2nd block of BERT-6 architecture pre-trained and tested on $W_S = W^{test} = 90{,}000$.

## 2.3 Asymmetric nearest-neighbor connections between tokens

The formation of token clusters assumes symmetric connections between tokens. This hard constraint is not always justified in a language, since if token 'A' is a strong or strongest match with token 'B', vice versa might be incorrect, as can be easily verified by exemplified synonyms appearing in thesaurus.com. In addition, indirectly connected tokens via several symmetric or asymmetric connections typically result in a drift meaning.

Using direct asymmetric connections of the confusion matrix $M_{Confus}$, the strong match of a given token with several others was analyzed. Each row of $M_{Confus}$ is normalized by its diagonal element and its K most confused tokens, the maximal K off-diagonal elements, are defined as Top-K (Fig. 11). It is evident that the Top-K tokens differ from the subset of tokens belonging to the cluster of that given token, as connections are not necessarily symmetric, and only nearest-neighbor connections were considered. Nevertheless, the first question is whether the Top-K represents a strong match to a given token, and the second is whether there is an overlap between the Top-K and the tokens belonging to the cluster of this given token (Figs. 7–10). A positive answer to these two questions will indicate the robustness of the identification of higher-order language structures by the pre-training process, independent of the selected measure type.

The exemplified Top-20 for selected tokens, some of which appeared in the $6^{th}$ transformer block clusters pre-trained over the entire Wikipedia dataset, were examined (Fig. 11). Typically, the leading Top-K has a strong match with the selected token and is in agreement with a large fraction of its token cluster (Figs. 4–5). For instance, Top-8 of 'may' are months or 'can', ten out of Top-12 of 'north' are directions, Top-7 of '15th' are small ordered numbers, Top-3 of 'biologist' are related to other research fields. Of course, there are unrelated Top-K tokens; however, they tyically do not appear in the first few tokens with the highest probabilities of confusion. In addition, because some of the selected tokens appear at a low frequency in the entire Wikipedia dataset, the lower Top-K tokens appear only once or twice and have negligible statistical significance. Similar Top-20 for the same tokens obtained from the $6^{th}$ transformer block pre-trained using $W_S = 90{,}000$ indicate similar results (not shown); however, with decreased strong match coherence among the tokens, as expected. In addition, strong Top-K matches with a given token significantly decrease in the $2^{nd}$ transformer block (not shown), indicating the progressive enhancement of higher-order language structures along the transformer blocks.

| Token | diego (1.000) | may (1.000) | north (1.000) | 15th (1.000) | designed (1.000) | small (1.000) | biologist (1.000) | electricity (1.000) |
|---|---|---|---|---|---|---|---|---|
| Decreasing top-k | francisco (0.159) | can (0.216) | south (0.173) | 17th (0.200) | developed (0.070) | large (0.067) | botanist (0.375) | energy (0.100) |
| | , (0.023) | february (0.072) | east (0.060) | 16th (0.175) | built (0.055) | two (0.014) | chemist (0.250) | heat (0.100) |
| | jose (0.023) | january (0.058) | west (0.037) | 14th (0.150) | used (0.051) | a (0.014) | physicist (0.250) | cooling (0.100) |
| | javier (0.011) | august (0.040) | the (0.017) | 19th (0.100) | intended (0.047) | larger (0.012) | academic (0.250) | infrastructure (0.100) |
| | michel (0.011) | april (0.038) | central (0.012) | 13th (0.100) | created (0.020) | new (0.012) | curator (0.125) | oil (0.100) |
| | jorge (0.011) | december (0.037) | northern (0.011) | 11th (0.100) | " (0.012) | the (0.012) | scientist (0.125) | railways (0.050) |
| | antonio (0.011) | march (0.033) | northwest (0.008) | 6th (0.075) | manufactured (0.008) | public (0.010) | teacher (0.125) | physics (0.050) |
| | pedro (0.011) | june (0.029) | northeast (0.007) | 1986 (0.050) | complete (0.008) | long (0.010) | author (0.125) | electric (0.050) |
| | ##do (0.011) | september (0.028) | in (0.006) | french (0.050) | Known (0.008) | power (0.010) | leader (0.125) | gdp (0.050) |
| | - | also (0.028) | southwest (0.005) | royal (0.050) | produced (0.008) | yellow (0.010) | ##ist (0.125) | science (0.050) |
| | - | , (0.023) | southern (0.005) | 1st (0.050) | invented (0.008) | other (0.007) | was (0.125) | own (0.050) |
| | - | july (0.020) | eastern (0.005) | ninth (0.050) | considered (0.008) | railway (0.007) | actor (0.125) | power (0.050) |
| | - | october (0.015) | of (0.005) | fourth (0.050) | described (0.008) | former (0.007) | man (0.125) | this (0.050) |
| | - | november (0.013) | new (0.005) | 12th (0.050) | able (0.008) | young (0.007) | and (0.125) | finance (0.050) |
| | - | to (0.012) | little (0.004) | second (0.050) | a (0.008) | private (0.007) | - | the (0.050) |
| | - | ##s (0.011) | southeast (0.003) | 2018 (0.025) | free (0.004) | big (0.007) | - | development (0.050) |
| | - | would (0.006) | , (0.003) | third (0.025) | worked (0.004) | single (0.007) | - | is (0.050) |
| | - | is (0.006) | main (0.003) | king (0.025) | research (0.004) | main (0.007) | - | : (0.050) |
| | - | people (0.005) | district (0.003) | former (0.025) | written (0.004) | book (0.007) | - | - |

**Fig. 11.** Selected tokens and their Top-20 tokens obtained from the confusion matrix, normalized each row by its maximal value (diagonal), using the 6$^{th}$ block of BERT-6 pre-trained on the entire Wikipedia dataset. The number below each token represents its normalized appearance frequency with respect to the diagonal (first token in each column).

## 2.4 Embedding layer – higher-order language structures

Most existing studies assume that language structures are mainly constructed during the pre-training process in the embedding layer[29-33]. These structures are expressed by the correlations between the vectors of rank $768$ assigned to each token[34-37]. The correctness of this statement is questioned below, where high-order language structures are further enhanced at the output layer using the intermediate pre-trained transformer blocks. The embedding layer does not stand alone, as the pre-training back-propagation mutually couples all blocks following the decision-making at the output layer. This coupling can be easily observed, where, for instance, the expected superior embedding layer of the pre-trained BERT-12[24] is used as a frozen embedding layer of BERT-6, and next the six transformer blocks are pre-trained. The <APT> is found to significantly decrease in comparison to the entire pre-trained BERT-6.

The correlations between pairs of embedding vectors of rank $768$ are typically measured by their cosine-similarity

$$\frac{\vec{V_i}\cdot\vec{V_j}}{\|V_i\|\|V_j\|} = \hat{V_i}\cdot\hat{V_j} \qquad (2)$$

where a value $1$ indicates that both vectors are identical up to a scalar pre-factor and $-1$ where the vectors are anti-parallel. The distribution of the cosine-similarity indicates a Gaussian shape centered around $0.45$, with continuous tails, where the peak at $1$ is attributed to the diagonal elements (Fig. 12) (similar results were obtained for the Euclidean measure distance among tokens). The continuous right tail indicates that, on average, there are no dominant tokens for a given token with high similarity, as observed in the distribution of the off-diagonal elements of the confusion matrix (Fig. 3). An additional fundamental difference is that the confusion matrix is asymmetric and practically highly diluted, whereas all elements of the symmetric cosine-similarity matrix are non-zero.

Finding high-language structures requires high dilution of the symmetric cosine-similarity matrix. Because clusters generated by the confusion matrix indicate that a token can have much more than $4$ strong match tokens (Figs. 5, 7-10), the cosine-similarity matrix is accordingly highly diluted. For each row of the similarity matrix the $q = 4$ maximal elements were set to one and the rest to zero. This binary similarity matrix is not necessarily symmetric, because if $j$ is one of the $q$ highest similar tokens of $i$, the token $i$ can be absent from the highest similar tokens of $j$. Nevertheless, the adjacency matrix is achieved using Eq. (1). Results for $q = 4$ indicate that the maximal leading clusters are $695, 693,$ and $396$, mixing incoherent tokens of words such as ‘attack’, ‘unable’, and ‘similar’ in a cluster, indicating a very low pre-train quality. To overcome this difficulty, $q$ was decreased to $3$, although the practicality of the low-$q$ limit is in question because the decay of the cosine-similarity for a given token is typically very slow (see below).

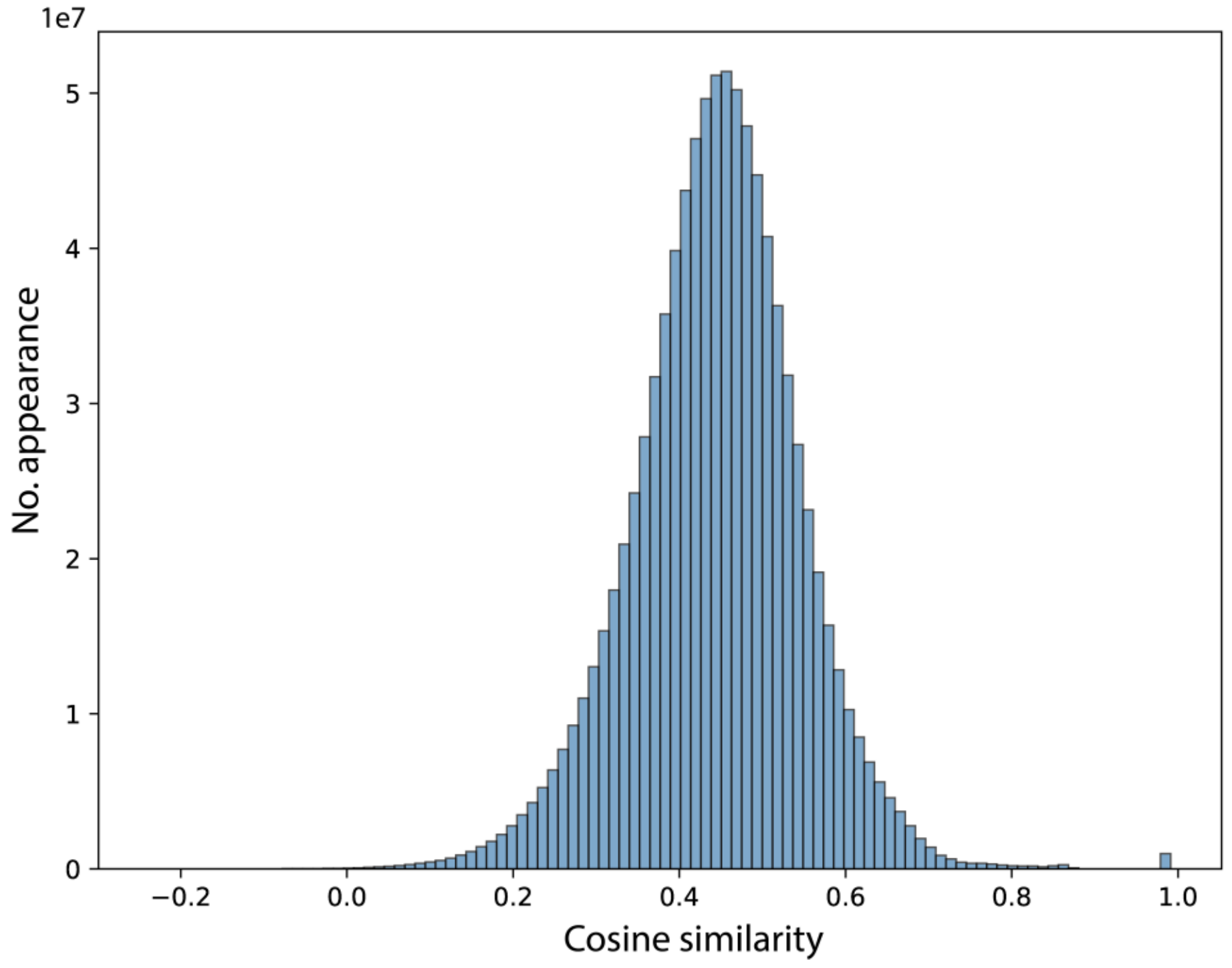

**Fig. 12.** Distribution of token–vector cosine-similarity pairs in the embedding layer of pre-trained BERT-6 over the entire Wikipedia dataset. The small rise at $1.0$ is attributed to the diagonal elements of the vectors multiplied by themselves.

The cluster size distribution for $q = 3$ indicates a long tail of clusters above $20$ where the maximum cluster size is $164$ (Table 3a). Some of the exemplified clusters include coherent tokens such as numbers from the 18$^{th}$ century (Fig. 13c) and a set of tokens that can replace each other while preserving the grammatical correctness of the sentence (Fig. 13d). However, a large fraction of the clusters clearly exhibits strong drift meaning such as ‘feelings’,

‘topographic’ and ‘financing’ (Fig. 13a), emerging from a few branches of the clusters, each composed of medium match tokens. Another type of clusters is composed of word inflections (Figs. 13b, 13g), which typically cannot represent grammatically correct exchange tokens in a sentence. The splitting of the $12$ months into a cluster of $7$ months (Fig. 13e) and a few other clusters (not shown) also indicates the incomplete gathering of similar objects (see Fig. 5b).

Decreasing the connectivity of each token further to $q = 2$, results in a smaller cluster distribution tail, with a maximum value of $44$ (Table 3b). The cluster shapes are now close to chains (Fig. 14) and have the same trends as those found for $q = 3$. Some numbers of the 20$^{th}$ century construct a cluster (Fig. 14d), several clusters demonstrate drift meaning (Fig. 14a, 14c, 14e, 14f, 14g), splitting the $12$ months into a few clusters (Fig. 14h), and a cluster containing inflections of a token (Figs. 14a, 14e, 14g).

Some of these features are a result of the sparsity of the clusters using $q = 2, 3$, in comparison with the clusters obtained by thresholding the confusion matrix (Fig. 3). However, increasing $q$ results in denser clusters and merges clusters with incoherent meanings into larger ones. It is important to note that although the number of classified clusters of size $1$ is large, as expected for very low connectivity, $q = 2, 3$ (Table 3), it is still much lower than the number obtained by the confusion matrix, which is $\sim 20{,}000$ (Table 2a).

The second proposed method to estimate the embedded high-language structures was the maximal K off-diagonal elements, Top-K, which are the K tokens most confused with the given one (Fig. 11). Using the cosine-similarity matrix, the Top-20 were exemplified (Fig. 15) for some tokens appearing in the clusters (Figs. 13 and 14). Similar features obtained for the clusters are observed for the Top-20, where for instance the leading Top-K for ‘electricity’ are ‘electrical’, ‘electrified, ‘electrically’ and ‘electrification’. In addition, the appearance frequency decays with K of the Top-K very slowly for many exemplified tokens, such that using a low $q(= 2$ or $3)$ is in question.

(a)

| Cluster Size | No. |
|---|---|
| 1 | 11,601 |
| 2 | 2,260 |
| 3 | 732 |
| 4 | 791 |
| 5 | 265 |
| 6 | 183 |
| 7 | 113 |
| 8 | 105 |
| 9 | 64 |
| 10 | 44 |
| 11 | 32 |
| 12 | 24 |
| 13 | 20 |
| 14 | 9 |
| 15 | 12 |
| 16 | 14 |
| 17 | 7 |
| 18 | 12 |
| 19,21 | 4 |
| 20 | 5 |
| 22,23,25, 27, 29, 40 | 2 |
| 24,26 | 3 |
| 28, 32, 45, 50,81,99, 164 | 1 |

(b)

| Cluster Size | No. |
|---|---|
| 1 | 13,397 |
| 2 | 3,127 |
| 3 | 1,469 |
| 4 | 757 |
| 5 | 228 |
| 6 | 102 |
| 7 | 41 |
| 8 | 15 |
| 9 | 12 |
| 10 | 2 |
| 11 | 3 |
| 12,13,15, 39,44 | 1 |

**Table 3.** (a) Distribution of clusters formed by using Top-3 of the cosine-similarity of the embedding layer obtained by BERT-6 pre-trained over the entire Wikipedia dataset. (b) Same as panel a but for Top-2.

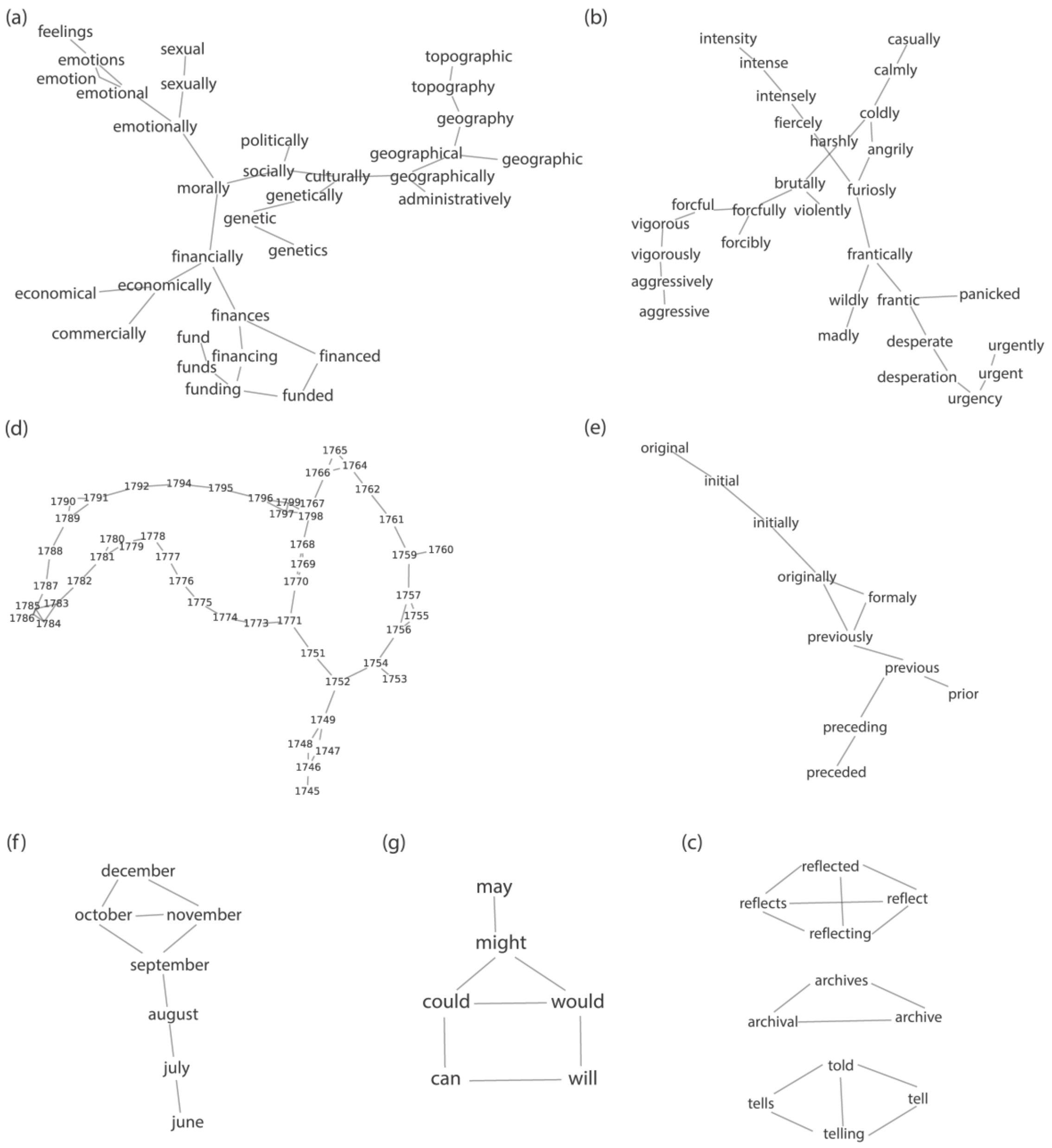

**Fig. 13.** Exemplified notable clusters generated by the cosine-similarity matrix of the embedding layer matrix of pre-trained BERT-6 over the entire Wikipedia dataset by applying Top-3 connections for each token.

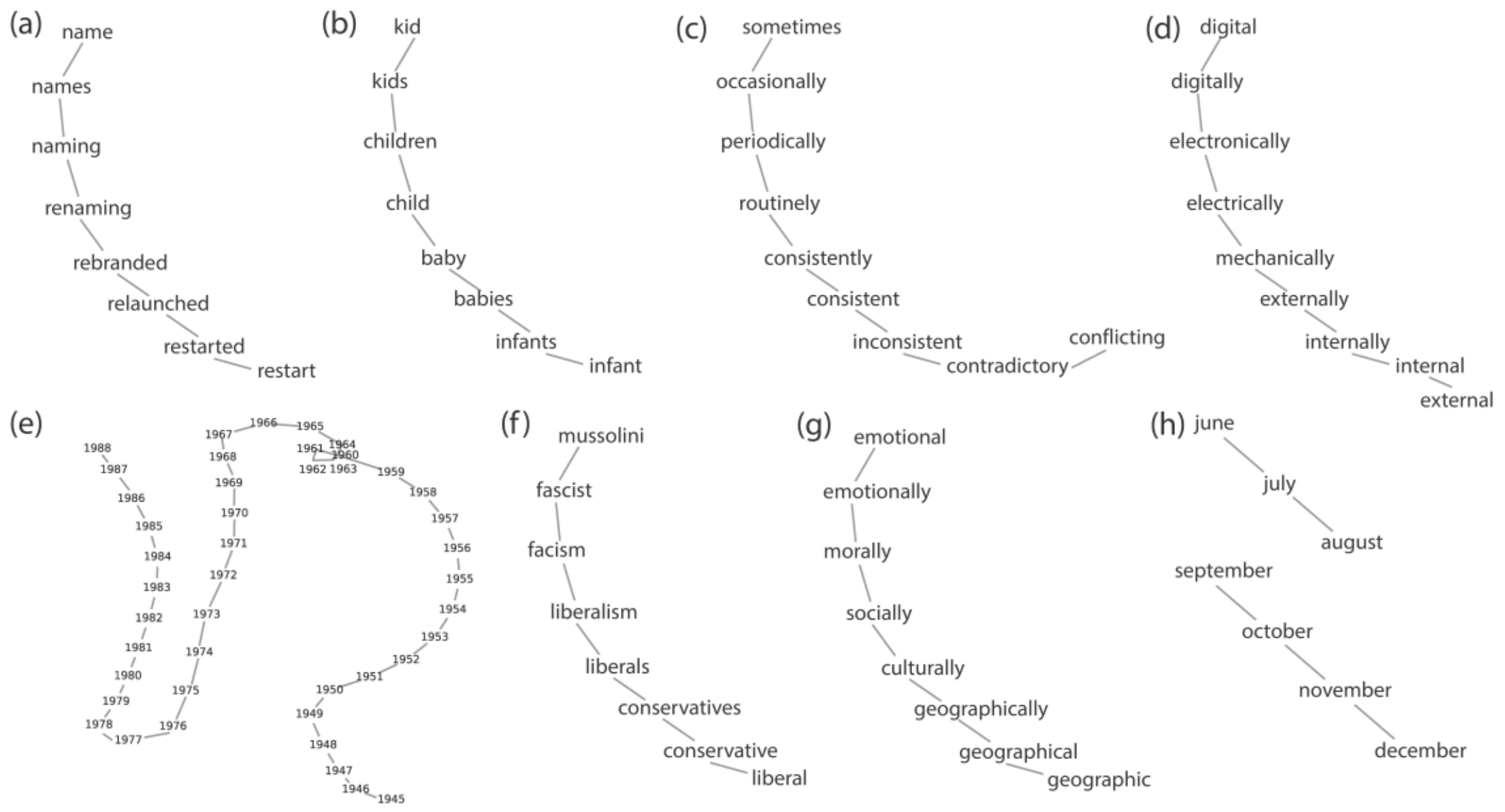

**Fig. 14.** Exemplified notable clusters generated by the cosine-similarity of the embedding layer matrix of pre-trained BERT-6 over the entire Wikipedia dataset by applying Top-2 connections for each token.

| Token | diego | may | north | 15th | designed | small | stated | biologist | electricity |
|---|---|---|---|---|---|---|---|---|---|
| Decreasing top-k | diego (1.0000) | may (1.0000) | north (1.0000) | 15th (1.0000) | designed (1.0000) | small (1.0000) | stated (1.0000) | biologist (1.0000) | electricity (1.0000) |
| | francisco (0.6461) | june (0.6619) | south (0.8015) | 14th (0.9027) | designing (0.6894) | large (0.7120) | stating (0.7179) | geologist (0.7415) | electrical (0.6225) |
| | bernardino (0.6320) | april (0.6378) | east (0.7117) | 16th (0.8911) | redesigned (0.6480) | smaller (0.6377) | commented (0.6340) | botanist (0.7367) | electrified (0.5829) |
| | [PAD] (0.5853) | might (0.6221) | west (0.6871) | 17th (0.8508) | designs (0.6341) | tiny (0.6235) | asserted (0.6322) | biology (0.7291) | electrically (0.5747) |
| | mateo (0.5836) | july (0.6151) | northern (0.5615) | 13th (0.8492) | design (0.6319) | larger (0.5276) | stipulated (0.6155) | physicist (0.7274) | electrification (0.5632) |
| | pablo (0.5659) | august (0.6065) | northeast (0.5492) | fifteenth (0.8331) | devised (0.5958) | little (0.5248) | remarked (0.6129) | naturalist (0.7079) | telecommunication (0.5542) |
| | ##二 (0.5631) | october (0.6011) | northwest (0.5406) | 12th (0.8314) | tailored (0.5834) | big (0.5040) | expressed (0.6101) | anthropologist (0.7064) | electric (0.5541) |
| | ##介 (0.5567) | march (0.5886) | southwest (0.4997) | 9th (0.8183) | redesign (0.5825) | sizable (0.4951) | specified (0.6090) | psychologists (0.6930) | generators (0.5427) |
| | ##宗 (0.5555) | november (0.5650) | southeast (0.4882) | 11th (0.8137) | designers (0.5757) | 259 (0.4923) | said (0.5860) | psychologist (0.6873) | waterfalls (0.5297) |
| | ก (0.5544) | september (0.5634) | northwards (0.4859) | 10th (0.8128) | engineered (0.5716) | 255 (0.4845) | reiterated (0.5856) | [PAD] (0.6865) | utilities (0.5289) |
| | 家 (0.5541) | february (0.5576) | northward (0.4781) | 23rd (0.8119) | designer (0.5710) | smallest (0.4845) | declared (0.5830) | zoology (0.6849) | sugarcane (0.5254) |
| | 外 (0.5536) | can (0.5500) | southern (0.4677) | 27th (0.8115) | constructed (0.5682) | 251 (0.4841) | disclosed (0.5825) | sociologist (0.6813) | energies (0.5238) |
| | ħ (0.5527) | january (0.5428) | nord (0.4514) | 29th (0.8050) | crafted (0.5644) | 324 (0.4840) | opined (0.5793) | theologian (0.6799) | outlawed (0.5211) |
| | ɳ (0.5515) | december (0.5373) | southward (0.4421) | 28th (0.7995) | configured (0.5615) | 284 (0.4830) | declares (0.5772) | scientists (0.6791) | empowered (0.5191) |
| | 地 (0.5513) | could (0.4982) | western (0.4312) | 24th (0.7993) | intended (0.5607) | 276 (0.4823) | affirmed (0.5753) | archaeologist (0.6753) | 1805 (0.5174) |
| | せ (0.5513) | will (0.4905) | nw (0.4299) | 25th (0.7967) | created (0.5590) | 305 (0.4820) | proclaimed (0.5744) | linguist (0.6750) | broadband (0.5172) |
| | matteo (0.5512) | would (0.4810) | 201 (0.4287) | 22nd (0.7964) | conceived (0.5552) | 307 (0.4816) | announced (0.5702) | neuroscience (0.6743) | 1835 (0.5169) |
| | ɯ (0.5509) | must (0.4315) | 218 (0.4281) | 30th (0.7953) | sculpted (0.5515) | 328 (0.4814) | commenting (0.5686) | psychiatrist (0.6714) | ㅊ (0.5166) |
| | ##三 (0.5501) | 189 (0.4146) | eastward (0.4279) | 26th (0.7935) | built (0.5506) | 306 (0.4804) | asserts (0.5685) | モ (0.6687) | aluminium (0.5161) |
| | い (0.5497) | 292 (0.4136) | 252 (0.4273) | 18th (0.7900) | formulated (0.5471) | 285 (0.4803) | decreed (0.5674) | ロ (0.6656) | electrons (0.5153) |

**Fig. 15.** Selected tokens and their Top-K tokens formed through the cosine-similarity matrix, using the embedding layer of BERT-6 pre-trained on the entire Wikipedia dataset. The number below each token represents its cosine-similarity values.

The proposed clustering method is a globally robust cluster-extraction technique. Other, more localized techniques exist with the aim of bias removal, which may provide better cluster extraction in successive clustering steps. One such method is the ABTT (All-But-The-Top)[32] method that removes the assumed unnecessary Top-r principal components (PCs) of embedding vectors, followed by Cross-domain Similarity Local Scaling (CSLS)[38] which accounts for local vector density while calculating cosine-similarity. Results indicate that such localized techniques could generate one huge cluster, where the overall small cluster formation, drift effect, and size are preserved. Another extension is the selection of an individual Top-q per token following its row statistics, which is similar to an individual threshold per token in the confusion matrix. Preliminary results indicate that the main characteristic features of the clusters are preserved under such local rules.

### 2.4 Confidence versus higher-order language structures

High-order language structures are embedded in pre-training; however, their indirect effect on the fine-tuning quality is unclear. Using the pre-trained BERT-6 over the entire Wikipedia dataset, fine-tuning over the FewRel dataset consisting of $64$ output labels resulted in $\sim 0.66$ accuracy.

The FewRel inputs consisted of a sequence of tokens with varying pre-training success, as measured by the APT. Hence, each FewRel input can be classified using the average APT over its tokens, $APT_{Ave}$, and the FewRel inputs can be grouped into bins following the $APT_{Ave}$ (Fig. 16a). For each bin, confidence is defined as

$$Confidence = \frac{N_{Correct}}{N_{Correct}+N_{Incorrect}} \quad (3)$$

where $N_{Correct}$ ($N_{Incorrect}$) represents the number of FewRel inputs belonging to a bin that are correctly (incorrectly) classified, respectively[5]. It is expected that the confidence of inputs with low $APT_{Ave}$ will be low because the ability to correctly understand and classify a FewRel input is screened by many incorrect tokens. Nevertheless, the results indicate no observed trend in $Confidence(APT_{Ave})$, which is practically independent of $APT_{Ave}$ (Fig. 16a). A possible explanation for this result is that, although a large fraction of the tokens' input is not correctly predicted by pre-training, the meaning of the FewRel inputs is preserved, as tokens are mainly replaced by their synonyms or strong-match tokens (Figs. 4–5, 7–10).

Similarly, the FewRel inputs were grouped using the average token appearance in the pre-training dataset. Because the APT, on average, increases with the appearance frequency of the token (Fig. 1), one might expect that confidence will increase with the average token appearance in the FewRel input. Nevertheless, the results indicate that confidence is independent of the token appearance (Fig. 16b); as a result that tokens are not randomly confused by other tokens, but mainly by strong-match tokens. Hence, $APT_{Ave}$ is an appropriate global order parameter for estimating the quality of the pre-training process; however, it does not directly reflect the quality of quantities, such as confidence. Nevertheless, the correlation between the success of the pre-training and fine-tuning processes is discussed in the next section.

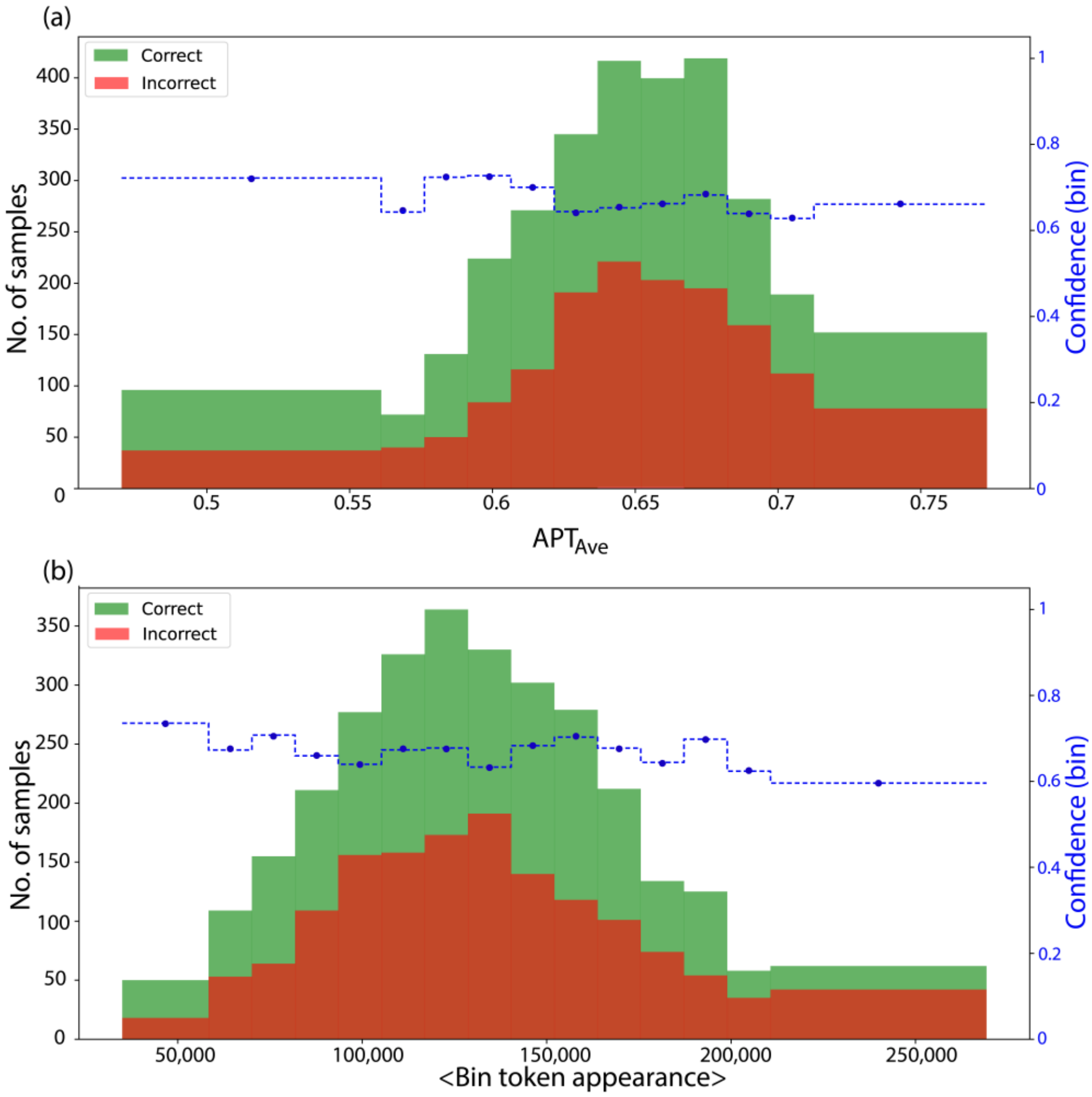

**Fig. 16.** (a) Number of FewRel inputs within a bin of $APT_{Ave}$, which are correctly (green) and incorrectly (red) classified, and the confidence per bin (blue dots connected by blue-dashed line) as defined in Eq. (3). (b) Similar to panel a but the bins are now grouped FewRel inputs whose average token appearance belongs to the bin's range.

### 2.5 Underlying Fine-tuning learning mechanism

The effectiveness of pre-training is evident, as a clear performance gap is observed between the pre-trained and non-pre-trained models across the classification tasks. This performance gap was recently found to increase with the size of the pre-training dataset and with a greater overlap between the tokens in the pre-training and classification datasets[5]. Here the correlation between the quality of the pre-training, measured by the global order parameter, <APT>, and the accuracy of the fine-tuning classification task, is exemplified by the FewRel and DBpedia datasets.

Understanding the underlying mechanism is based on the single-nodal performance (SNP)[39]. It quantifies the functionality of each of the $768$ output nodes of the $m^{th}$ transformer block which are FC to an output layer consisting of $N_{Lables}$ nodes. This FC layer is trained to minimize the loss function, whereas the fine-tuning-trained weights of all $m$ blocks and the embedding layer were kept frozen (unchanged). Next, all the weights of the FC layer were silenced, except for the specific $N_{labels}$ weights that emerge from a single node. At this point, the validation dataset is presented and preprocessed by the first $m$ transformer blocks, while influencing the output units only through the small aperture of one node, generating an $N_{Labels} \times N_{Labels}$ value matrix. Each matrix element $(i, j)$ represents the average field generated on the output unit $j$ by validation inputs with label $i$ normalized by the maximum matrix element. A Boolean-clipped matrix was then derived by applying a threshold followed by a permutation to form diagonal clusters[40, 41]. Elements located outside these diagonal clusters above the threshold were classified as noise $n$. The results demonstrate (Table 4) that each node essentially contains on average less than two clusters, $N_C$, where the average cluster size approaches one, $C_S$, with additional above-threshold elements out of the diagonal cluster identified as noise elements, $n$. One can show that the signal-to-noise ratio (SNR) is given by

$$SNR = \frac{768 \cdot \frac{Diag}{N_{Labels}}}{768 \cdot \frac{n}{N_{labels}^2}} = N_{Labels} \cdot \frac{Diag}{n} \quad (4)$$

where the l. h. s. numerator (denominator) represents the signal (noise) that increases along the blocks.

| Block | Acc. | Diag | $N_C$ | $C_S$ | $n$ | SNR |
|---|---|---|---|---|---|---|
| 6 | 0.64 | 1.49 | 1.34 | 1.11 | 12.5 | 7.63 |
| 5 | 0.63 | 1.57 | 1.32 | 1.19 | 16.7 | 6.01 |
| 4 | 0.60 | 1.96 | 1.40 | 1.41 | 30.9 | 4.05 |
| 3 | 0.59 | 2.49 | 1.55 | 1.61 | 53.7 | 2.97 |
| 2 | 0.54 | 2.62 | 1.53 | 1.71 | 70.7 | 2.38 |
| 1 | 0.47 | 3.46 | 1.59 | 2.17 | 151.2 | 1.46 |

**Table 4.** Statistical properties of the $768$ SNP using threshold $0.6$ measured at the end of each block of the pre-trained BERT-6 over the entire Wikipedia dataset and next fine-tuning on the FewRel dataset, $N_{Labels} = 64$. The table reports, FewRel accuracy, Acc., average number of clusters per SNP matrix, $N_c$, average cluster size, $C_s$, average number of diagonal elements per SNP matrix, $Diag$, average noise per SNP matrix, $n$, and the signal-to-noise ratio, SNR (Eq. (4)).

Similar trends were obtained for the DBpedia dataset, which consists of $14$ output labels, with $40{,}000/5{,}000$ train/test instances per label, comprising $28{,}621$ unique tokens. When using the full dataset, the accuracy with and without pre-training was $\sim 0.99$, with a minimal gap of $< 0.01$[42], and the difference between the obtained accuracies of blocks $1$ and $6$ is only $0.01$. To observe meaningful gaps in the accuracy along the transformer blocks, the dataset size was significantly reduced to $100$ train/test instances per label (Table 5). The trends of the SNP are similar to those of the FewRel case (Table 4), where the accuracy increases along the transformer blocks, $N_C$ and $C_S$ decrease towards unity, $n$ decreases by a factor ten from the first to the 6$^{th}$ transformer block, and the SNR increases along the blocks. The increase in the SNR along the architecture (Tables 4–5) forms the underlying fine-tuning learning mechanism.

| Block | Acc. | Diag | $N_C$ | $C_S$ | $n$ | SNR |
|---|---|---|---|---|---|---|
| 6 | 0.979 | 1.3 | 1.2 | 1.1 | 2.56 | 6.86 |
| 5 | 0.978 | 1.4 | 1.2 | 1.2 | 4.19 | 4.44 |
| 4 | 0.970 | 1.5 | 1.2 | 1.2 | 6.24 | 3.20 |
| 3 | 0.947 | 1.7 | 1.2 | 1.4 | 9.70 | 2.26 |
| 2 | 0.927 | 1.8 | 1.2 | 1.5 | 11.7 | 2.03 |
| 1 | 0.828 | 2.3 | 1.3 | 1.8 | 20.6 | 1.45 |

**Table 5.** Statistical properties of the $768$ SNP using threshold $0.6$ measured at the end of each block of the pre-trained BERT-6 over the entire Wikipedia dataset and next fine-tuned on the DBpedia dataset, $N_{Labels} = 14$, using $100$ train/test instances per label. Notations are the same as in Table 4.

The understanding of the functionality of the multi-head attention (MHA)[4, 18] of each transformer block and each of its $12$ heads is realized using the following procedure based on a single matrix value representing the single-head performance (SHP)[17, 18]. The first $m-1$ transformer blocks and the QKV attention of the $m^{th}$ transformer encoder block in the pre-trained BERT-6 architecture were kept frozen. The $768$ outputs of the scaled dot-product attention were FC to the output layer consisting of $N_{Labels} = 64$ units, representing the FewRel labels. The FC layer was trained using a validation dataset to minimize the loss function, similar to the procedure at the end of the block (Table 4), and the accuracy of the entire MHA was estimated.

The functionality of each of the $12$ heads, $H(m)$ $(m = 1, 2, \ldots, 12)$, was estimated by silencing all input nodes to the trained FC layer, except for the $64$ nodes belonging to the examined head. Consequently, the $64$ output units representing the FewRel labels were influenced only by the $64$ $(= \frac{768}{12})$ nodes of that head. The validation dataset was then propagated through the first $m-1$ pre-trained transformer blocks, as well as through the silenced QKV attention of the $m^{th}$ block, generating a $64 \times 64$ value matrix. A Boolean-clipped matrix was then derived by applying a threshold followed by a permutation to form diagonal clusters. Elements above the threshold but located outside these diagonal clusters were classified as noise $n$.

Results of the 6th transformer block indicate that each SHP matrix comprises multiple clusters of approximately one unit size, which fluctuates among the 12 heads, with an average $N_C \sim 20$ unit clusters per head (Table 6). These unit cluster elements indicate that, on average, an input is correctly classified unless it competes with the existing noise elements, $n$; $\sim 6$ noise elements distributed on a $64 \times 64$ matrix (Table 6). Extension of these results to the QKV attention for all transformer blocks ($\geq 2$), indicates that the $C_S$ approaches unity for all blocks and noise remains relatively small per matrix element (Table 7). Attention accuracy increased along the blocks and was mainly attributed to an increase in the diagonal, which was proportional to the signal (similar to Eq. (4)).

| No. Head | Diag | $N_C$ | $C_S$ | $n$ |
|---|---|---|---|---|
| 1 | 23 | 23 | 1.00 | 3 |
| 2 | 29 | 28 | 1.04 | 4 |
| 3 | 12 | 10 | 1.20 | 16 |
| 4 | 33 | 32 | 1.03 | 10 |
| 5 | 21 | 21 | 1.00 | 3 |
| 6 | 22 | 22 | 1.00 | 4 |
| 7 | 23 | 23 | 1.00 | 2 |
| 8 | 25 | 25 | 1.00 | 0 |
| 9 | 15 | 14 | 1.07 | 3 |
| 10 | 11 | 10 | 1.10 | 22 |
| 11 | 7 | 7 | 1.00 | 1 |
| 12 | 27 | 26 | 1.04 | 3 |
| Ave | 20.7 | 20.1 | 1.04 | 5.92 |

**Table 6.** Statistical properties of the 12 SHP using threshold 0.6 measured at the attention layer of the 6th block of BERT-6 pre-trained over the entire Wikipedia dataset and then fine-tuned on the FewRel dataset[27, 28]. The table reports number of clusters per SHP matrix, $N_c$, their average cluster size, $C_s$, number of diagonal elements per SHP matrix, $Diag$, noise per SHP matrix, $n$. The last row reports the average over the 12 SHP matrices.

| Block | Acc. | Diag | $N_C$ | $C_S$ | $n$ |
|---|---|---|---|---|---|
| 6 | 0.64 | 20.7 | 20.1 | 1.04 | 5.92 |
| 5 | 0.63 | 12.6 | 11.7 | 1.12 | 3.92 |
| 4 | 0.60 | 8.1 | 6.9 | 1.28 | 2.25 |
| 3 | 0.59 | 6.2 | 5.4 | 1.02 | 11.42 |
| 2 | 0.55 | 6.7 | 5.9 | 1.13 | 13.25 |

**Table 7.** Average statistical properties of the QKV attention for blocks $\geq 2$ based on the statistical properties of their $12$ SHP using threshold $0.6$ (similar to Table 6). The table reports average number of clusters per SHP matrix, $N_c$, average cluster size, $C_s$, average number of diagonal elements per SHP matrix, $Diag$, average noise per SHP matrix, $n$. Reported results are for pre-trained BERT-6 over the entire Wikipedia dataset and then fine-tuned on the FewRel dataset.

Results indicate that enhanced fine-tuning accuracy (Tables 4–5) requires enhanced SNR (Eq. 4)). This observation was based on the averaged accuracy over the $64$ FewRel labels and the $14$ DBpedia labels but can also be extended to the accuracy per label. The average label cluster appearance in the SHP clusters, $N$, fluctuated significantly among the $64$ labels (Fig. 17). Hence, the signal and SNR (Eq. (4)) are expected to fluctuate among the labels. The results indicated a positive correlation between the label cluster appearance and fine-tuning accuracy. Increasing the signal enhanced the average accuracy of the labels (Fig. 18).

The underlying learning mechanism of the FewRel and DBpedia classification based on pre-trained BERT-6 over the Wikipedia dataset is similar to the mechanism obtained for the classification of CIFAR[43] using convolutional deep neural networks[44-46] and vision transformer architectures[20]. Hence, the results hint at its universality[40, 47], independent of the details of the classification tasks, pre-training, and feedforward architectures.

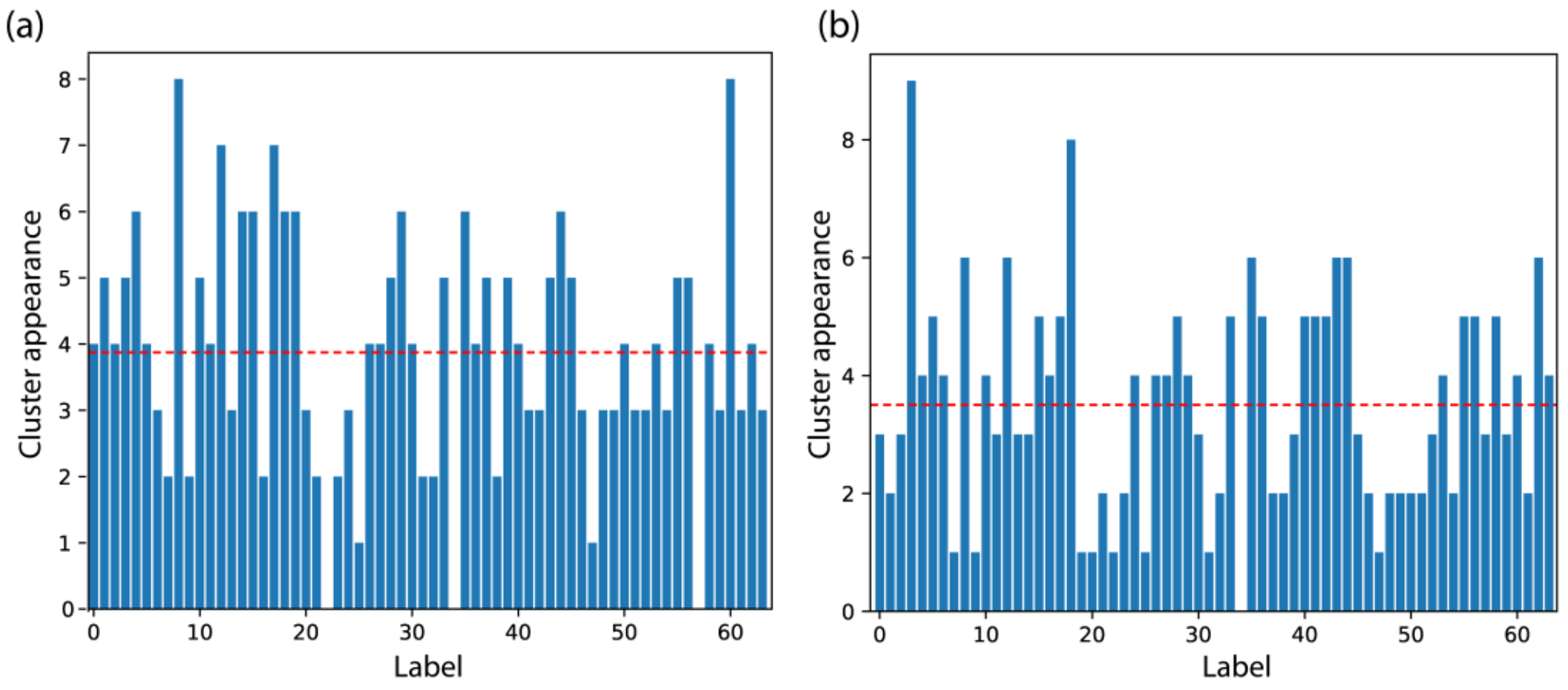

**Fig. 17.** (a) Distribution of label appearance in the diagonal clusters of the $12$ SHP of the 6th transformer block of BERT-6 pre-trained on $W_S = 90{,}000$. (b) Same as in panel a but pre-trained on $W_S = 40{,}000$.

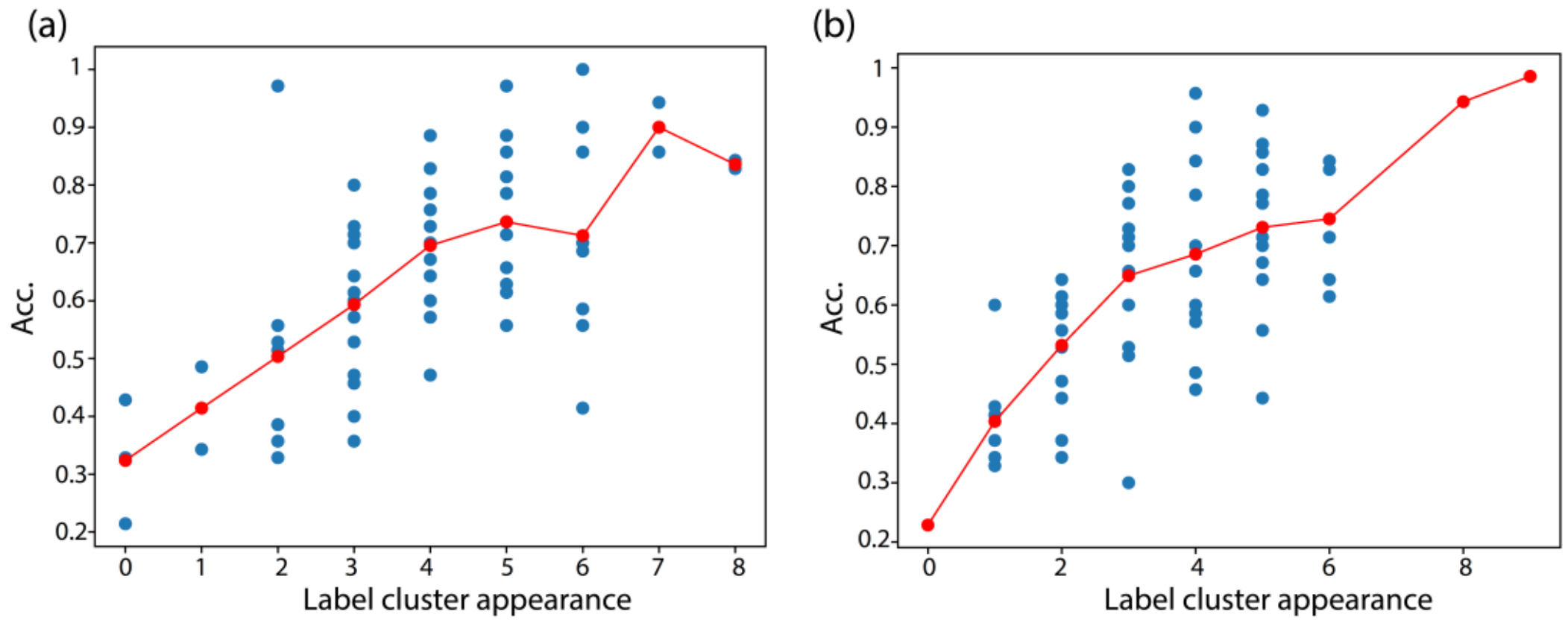

**Fig. 18.** (a) Accuracy (Acc.) for each of the 64 FewRel labels (blue circles) versus its number of appearances in the diagonal clusters of the $12$ SHP of the 6th transformer block of BERT-6 pre-trained on $W_S = 90{,}000$. (b) Same as panel a but pre-trained on $W_S = 40{,}000$.

## 3. Discussion

One of the noteworthy results is that pre-training breaks the symmetry among tokens and groups them into finite, small clusters of strong match tokens. This type of cluster formation is inferred from the token confusion matrix; when the predicted pre-trained selected token is incorrect, this token is replaced with a strong match token with a high probability. Assuming mutually symmetric relationships between pairs of tokens, the adjacency matrix indicates the formation of clusters, similar to the percolation phenomenon[25, 26]; however, existing connections are not selected randomly. This exemplifies the generation of high-order language structures by pre-training, which is sharpened along the transformer blocks toward the output layer, although the cost function of training is based solely on identifying a single token.

The robustness of such supreme language learning capabilities is supported by the asymmetric confusion matrix direct measure using nearest-neighbor Top-K connections. The maximum small number of K off-diagonal elements, the K most confused tokens with the given one, were typically found to be with its strong meaning match tokens.

An alternative viewpoint assumes that the main high-order language features are embedded in the cosine-similarity among pre-trained long embedding layer vectors representing each token[29-37]. The large matrix embedding layer space and large vector rank are attractive to comprise complex correlations among tokens, where the output layer is only a vector, devoting one output unit per token. Consequently, higher-order structures can be expected to be expressed by similarities among the $768$ rank vectors assigned to each token.

However, the space of the suggested confusion matrix is larger than that of the embedding layer. Moreover, the embedding layer is not a stand-alone entity; the pre-training coupled all weights of the embedding layer and the blocks following decision making at the output layer (Table 1). A simple experiment supports this claim where the expected superior embedding layer of pre-trained BERT-12, which leads to higher accuracy, is used as a frozen embedding layer of BERT-6, and next the six transformer blocks are pre-trained. It is found that the quality of such pre-training, measured by <APT>, is significantly reduced in comparison to the entire pre-trained BERT-6. A direct measure of strong match tokens using a symmetric and FC similarity matrix requires high dilution to prevent a large cluster containing a large fraction of the tokens. The criterion for such dilution is artificial because the cosine-similarity distribution is continuous. Additionally, its realization results in the aggregation of far- or drift meaning tokens and tokens that cannot grammatically replace each other in a sentence. Nonetheless, although high-order language structures are highly expressed in the output layer, as measured by the confusion matrix, the pre-trained embedding layer serves as their significant initiative learning step.

We conclude with a perspective on two interesting future directions—that is, theoretical and practical concepts. The theoretical approach concentrates on the entropy of a language, whose general definition, the asymptotic number of grammatically correct large texts, is an open question. Nevertheless, the presented work raises the question of how to estimate the entropy of strong match tokens. The case in which all the $APT$ increase indicates a lower entropy with a better definition or identification of the text. However, the comparison between more structured scenarios remains unclear. For example, a cluster of two strong match tokens with $APT = 0.6$ for each, versus $APT = 0.75$ for each token in two clusters of size one. The preferable scenario is not necessarily related to a pure entropic measure but depends on the fine-tuning task.

In this study, a strong match among single tokens was demonstrated using pre-trained architectures comprising a single weight connecting each two nodes and an assigned vector to each token in the embedding layer. In image classification tasks it was found that input crosses, representing higher-order correlations among input nodes can enhance accuracy, as well as the hyper-weights connecting several nodes between the deep architecture layers[48-52]. These findings suggest the generalization of pre-trained architectures to include input crosses, coupling more than one embedding vector, or hyper-weights. This can lead to higher-order language structures in which a strong match between tokens can be generalized to a strong match between phrases or several consecutive tokens. The implementation of input crosses and hyper-weights increases the feedforward and backpropagation complexity considerably. However, their numerous biological realities[53] are the byproducts of the nonlinear adaptive dendritic hardware implementing imitating dendritic learning[53-56]. These experimental findings on brain activity hint at the importance of including a similar feature in NLP.

Finally, the results were based on small-scale simulations and limited dataset diversity, and broader validation across different NLP domains should be examined in the future.

**Acknowledgements**

The work is supported by the Israel Science Foundation [grant number 346/22].

## Appendix

*1. Dataset and preprocessing*

The datasets used in this study are DBpedia[57], FewRel[58] and Wikipedia[6]. Each dataset was tokenized using the BERT tokenizer from the HuggingFace transformers library[10], specifically the bert-base-uncased variant[24], which converts raw text into $30{,}522$ token IDs. Tokenization was performed using the following configuration: truncation to a maximum length of $128$ tokens and padding to the same length.

*2. Optimization*

The CrossEntropyLoss[59] function was selected for the classification task and minimized using the stochastic gradient descent algorithm[60, 61] and the AdamW optimizer[62] was used. The maximal accuracy was determined by searching through the hyper-parameters (see below). The L2 regularization method[63] was applied.

To pre-train our models, we employed a Masked Language Modeling (MLM) objective similar to that used in the original BERT architecture[24]. Pre-training was performed on a filtered Wikipedia-derived dataset. The MLM procedure followed the standard masking strategy[10]: $15\%$ of tokens were selected for masking, of which $80\%$ were replaced with [MASK], $10\%$ with random tokens, and $10\%$ remained unchanged. Special tokens (e.g., [PAD], [CLS], [SEP]) were excluded from masking.

*3. Hyper-parameters*

The hyper-parameters $\eta$ (learning rate) and $\alpha$ (L2 regularization) were optimized for offline learning, using a mini-batch size of $32$ inputs. The learning-rate decay schedule was also optimized. For all the simulations, a linear scheduler was applied using the HuggingFaceutility[10], with zero warm-up steps and a total number of training steps equal to the number of epochs. This schedule gradually decays the learning rate from its initial value to zero in a linear fashion throughout training, which helps stabilize convergence[64]. The pre-training models in all the simulations were trained for $50$ epochs with $\eta = 5.5e-5, \alpha = 1e-2$ and a linear scheduler was applied.

For the statistics involving the mechanism of NLP pre-training, we utilized the pre-trained DistilBERT model of BERT-6 using the HuggingFace transformers library[65, 66].

For Table 4 and Figs 17-18 we pre-trained a BERT-6 model on a custom Wikipedia-based dataset, consisting of either $90{,}000$ or $40{,}000$ paragraphs. Following pre-training, the model was fine-tuned on the FewRel dataset with $\eta = 5e-5, \alpha = 5e-3$. The model was then cut at individual layers and an individual output layer was trained with different hyper-parameters for each layer displayed below in Tables 8-10. Hyper-parameters were optimized to reach maximal accuracy in each layer.

| FewRel hyper-parameters with $W_S = 90{,}000$ | | |
|---|---|---|
| Layer | $\eta$ | $\alpha$ |
| 6 | 5e-5 | 8e-6 |
| 5 | 1e-4 | 8e-6 |
| 4 | 1e-4 | 8e-6 |
| 3 | 5e-5 | 8e-6 |
| 2 | 1e-4 | 8e-6 |
| 1 | 5e-2 | 8e-6 |

**Table 8.** Hyper-parameters for FC output layer of BERT-6 fine-tuned on FewRel with $W_S = 90{,}000$.

| FewRel hyper-parameters with $W_S = 40{,}000$ | | |
|---|---|---|
| Layer | $\eta$ | $\alpha$ |
| 6 | 1e-4 | 8e-6 |
| 5 | 1e-4 | 8e-6 |
| 4 | 1e-4 | 8e-6 |
| 3 | 1e-4 | 8e-6 |
| 2 | 1e-4 | 8e-6 |
| 1 | 2e-2 | 8e-6 |

**Table 9.** Hyper-parameters for FC output layer of BERT-6 fine-tuned on FewRel with $W_S = 40{,}000$.

Similarly, for Table 5 the pre-trained BERT-6 with $W_S = 90{,}000$ was fine-tuned on DBpedia using $100$ instances for training and testing. The fine-tuning hyper-parameters are displayed below in Table 10.

| DBpedia hyper-parameters | | |
|---|---|---|
| Layer | $\eta$ | $\alpha$ |
| 6 | 5e-5 | 8e-6 |
| 5 | 1e-4 | 8e-6 |
| 4 | 1e-4 | 8e-6 |
| 3 | 5e-5 | 8e-6 |
| 2 | 1e-4 | 8e-6 |
| 1 | 5e-2 | 8e-6 |

**Table 10.** Hyper-parameters for FC output layer of BERT-6 fine-tuned on DBpedia.

*4. Cluster formation and Percolation*

The confusion matrix was formed by iterating the Wikipedia dataset with $W_S = 90{,}000$, masking a single token each time with $80\%$ chance of masking, $10\%$ chance replacement with a random token and $10\%$ no change at all, over the pre-trained BERT-6 architecture. For each token which was masked, $1$ was added in the $30{,}522 \times 30{,}522$ confusion matrix corresponding, where each cell $(i, j)$ represents the number of times output $j$ was selected for input $i$. Rows which had maximal off-diagonal elements were then diluted from the matrix due to misrepresentation of bad data, note that this does not correlate to high accuracy.

After the confusion matrix is created, each non-zeros row was normalized by the diagonal element so that each cell was between $0$ and $1$. A threshold was then applied, marking $0$ all below threshold elements and $1$ above. The clipped matrix was then element-wise multiplied by its transpose, creating a symmetrical matrix which conserves only dual connections among tokens (both $(i, j)$ and $(j, i)$ are $1$).

The clusters are then formed through percolation, where each cluster consists at first of a non-zero token row. All tokens that are “connected” (i.e. have $1$ in that token’s row) are then added to the cluster. The process is repeated for every newly added token until no connections to new tokens exist.

*5. Cosine-similarity matrix*

The cosine-similarity matrix was created by multiplying each normalized embedded token vector of $768$ by all the other normalized token vectors creating a $30{,}522 \times 30{,}522$ matrix with cosine-similarity values representing the distance between different normalized vectors with a diagonal of $1$. The matrix was then clipped using the Top-K method, where each the Top-K non-diagonal elements where set to $1$ and the others $0$. After the clipping, the clusters were formed using the percolation method.

*6. Statistics*

Statistics for all results were obtained using at least five samples and the standard division was around $1\%$ for all the results.

*7. Hardware and software*

We used Google Colab Pro and its available GPUs. We used Pytorch for all the programming processes.